\documentclass{article}

\usepackage[table,xcdraw]{xcolor}
\usepackage[preprint]{neurips_2025}

\usepackage[utf8]{inputenc} 
\usepackage[T1]{fontenc}    
\usepackage{url}            
\usepackage{booktabs}       
\usepackage{amsfonts}       
\usepackage{nicefrac}       
\usepackage{microtype}      

\usepackage{microtype}
\usepackage{graphicx}
\usepackage{subfigure}
\usepackage{multirow}   
\usepackage{graphicx}   
\usepackage{array}      
\usepackage{longtable}  
\usepackage{pifont}
\usepackage[most]{tcolorbox}
\usepackage{wrapfig} 
\usepackage{float}
\usepackage{stfloats}

\usepackage{bm}
\usepackage{graphicx}  
\usepackage{titletoc}  
\usepackage{lipsum}    
\usepackage{subcaption} 
\usepackage{enumitem}
\usepackage{minitoc} 
\usepackage{amsmath}
\usepackage{amssymb}
\usepackage{mathtools}
\usepackage{amsthm}
\definecolor{Gray}{rgb}{0.5, 0.5, 0.5}
\usepackage{natbib}
\usepackage{algorithm2e}     
\usepackage{graphicx}
\usepackage{array}   
\usepackage{rotating} 
\usepackage[textsize=tiny]{todonotes}
\usepackage{fontawesome5}    
\usepackage{caption}
\usepackage{pst-text}
\usepackage{graphicx}
\definecolor{edit_haohh}{rgb}{0,0.7,0}  

\definecolor{lightpink}{rgb}{1.0, 0.75, 0.8} 
\definecolor{lightyellow}{rgb}{1, 0.8, 0.02} 
\definecolor{lightgreen}{rgb}{0.651, 0.639, 0.008} 
\definecolor{codeblue}{rgb}{0.4,0.4,0.7}  

\newcommand{\TFicon}[1]{%
  \ifnum\pdfstrcmp{#1}{F}=0
    {\textcolor{codeblue}{\faSnowflake}}
  \else
    {\textcolor{pink}{\faFire}}
  \fi
}

\title{%
  \resizebox{0.18\textwidth}{!}{%
    \begin{tikzpicture}[scale=1,transform shape,baseline]
      \node[anchor=base,text=cyan,opacity=1] at (0.1,-0.2) {\textbf{\Huge CoNav}};
      \node[anchor=base,text=codeblue] at (0,-0.1) {\textbf{\Huge CoNav}};
    \end{tikzpicture}%
  }%
: Collaborative Cross-Modal Reasoning for Embodied Navigation
}

\author{
 Haihong Hao$^{1}$, ~ Mingfei Han$^{2}$, ~Changlin Li$^{3}$, ~Zhihui Li$^{1}$, ~Xiaojun Chang$^{1}$
\and
{
\small
$^1$ University of Science and Technology of China \;  
$^2$ MBZUAI \;
$^3$ Stanford University\;
}
}

\pdfoutput=1

\begin{document}

\maketitle
\vspace{-15pt}
\begin{center}
\footnotesize{
Project Page: \url{https://oceanhao.github.io/CoNav/}}
\end{center}
\vspace{30pt}
\begin{abstract}
Embodied navigation demands comprehensive scene understanding and precise spatial reasoning. While image–text models excel at interpreting pixel‐level color and lighting cues, 3D–text models capture volumetric structure and spatial relationships. However, unified fusion approaches that jointly fuse 2D images, 3D point clouds, and textual instructions 
face challenges in limited availability of triple-modality data and difficulty resolving conflicting beliefs among modalities.
In this work, we introduce \textbf{CoNav}, a collaborative cross-modal reasoning framework where a pretrained 3D–text model explicitly guides a image-text navigation agent by providing structured spatial-semantic knowledge to resolve ambiguities during navigation.
Specifically, we introduce Cross-Modal Belief Alignment, which operationalizes this cross-modal guidance by simply sharing textual hypotheses from the 3D–text model to the navigation agent. 
Through lightweight fine-tuning on a small 2D–3D–text corpus, the navigation agent learns to integrate visual cues with spatial-semantic knowledge derived from 3D-text model, enabling effective reasoning in Embodied navigation.
CoNav achieves significant improvements on four standard embodied navigation benchmarks (R2R, CVDN, REVERIE, SOON) and two spatial reasoning benchmarks (ScanQA, SQA3D). Moreover, under close navigation Success Rate, CoNav often generates shorter paths compared to other methods (as measured by SPL), showcasing the potential and challenges of fusing data from different modalities in embodied navigation.
\end{abstract}
\section{Introduction}
\label{sec:intro}
Embodied AI aims to enable agents to interact with the physical 3D world. One key challenge here is embodied navigation task \cite{anderson2018evaluation, savva2019habitat,ai2thor}. In this task, the navigation agent needs to move through a 3D environment based on user natural‐language instructions. In the past, most embodied navigation methods relied on only 2D visual cues \cite{chen2022duet,li2022bevformer,zhou2025navgpt,wang2024vision,han2024roomtour3d} by sampling from real-world scenes, which limited the agents' ability to understand spatial-semantic knowledge. 

More recently, 3D Large Language Models (3D-LLMs) have emerged, enabling agents to understand spatial relationships by integrating 3D cues like multiview\cite{10.1109/ICRA48506.2021.9561631}, bird's-eye view \cite{liu2023bird}, voxel \cite{liu2024volumetric}, point cloud \cite{3D-LLM, xu2025pointllm, yang20243d} etc. into LLMs. Despite this progress, applying 3D-LLMs to practical tasks like embodied navigation remains challenging. Recent fusion-based navigation methods have attempted to fuse 2D images, 3D point clouds, and text instructions into unified models. For example, LEO \cite{huang2023embodied} employs three separate encoders to process 2D images, 3D point clouds, and text instructions. The output features from these encoders are then treated as tokens and fed into a single LLM, enabling cross-modal reasoning. LEO trained on 1539 K data pairs and fuse multimodal data, tackle tasks such as 3D scene captioning, embodied navigation and manipulation. Similarly, VER \cite{liu2024volumetric} uses two encoders to separately process text instructions and multi-view observations. The resulting features are then fused and passed through a shared multi-layer transformer to perform cross-modal reasoning. These methods integrate data from different modalities at feature-level within a single unified architecture.

\begin{figure*}[t]
    \centering
    \includegraphics[width=1\textwidth]{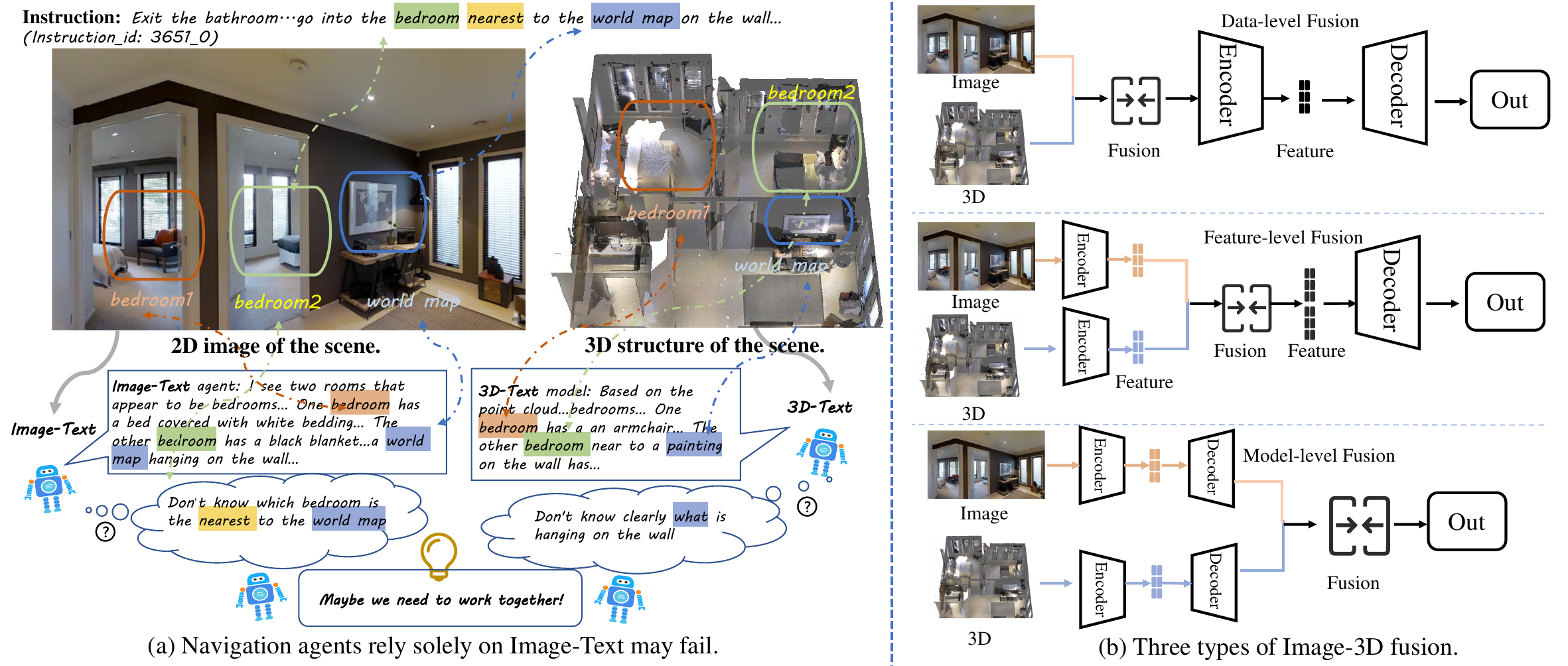} 
    \caption{\small (a) Image–text navigation agents rely on only visual cues may fail when dealing with tasks involving spatial distance. Although 3D–text models excel in spatial reasoning, they are less sensitive to textures (e.g., map and painting on the wall). Embodied navigation often demands both visual cues and spatial geometry for reasoning. In the scene of Figure, the agent need to use visual cues to locate a world map and then apply spatial distance reasoning to determine which bedroom is nearest to that map. (b) Fusion of image and 3D can be achieved through three methods. Past unified fusion approaches often employed multimodal fusion at the feature level to complete cross-modal reasoning. By enabling text-based information sharing at the model level, CoNav achieves higher-level cross-modal reasoning.}
    \label{fig:fig1}
\end{figure*}

Fusion-based navigation methods face two drawbacks. First, they require large corpora of 2D–3D–text triples—yet high-quality datasets typically exist as image–text \cite{chen2015microsoft, changpinyo2021cc12m,deng2009imagenet, kakaobrain2022coyo-700m} or 3D–text pairs \cite{Cap3D,Objaverse,PointCLIPV2,xu2025pointllm}, making comprehensive triple-modality collection difficult. Second, monolithic fusion at feature-level as shown in Figure \ref{fig:fig1} may obscure each modality’s unique characteristics and eliminates cross-verification by different modalities.
Moreover, prior research \cite{yuan2023mlink,trirat2024automl} has shown that, under a fixed parameter budget, specialized modular processing often yields better performance than a single monolithic model.

In fact, the image–text and 3D–text modalities each offer distinct strengths, as illustrated in Figure~\ref{fig:fig1}. Image-text models rely on pixel-level color and illumination changes to infer boundaries and semantics \cite{radford2021learning, jia2021scaling}, 3D-text models prioritize spatial geometry and volumetric structure \cite{chen2020scanrefer, chen2021scan2cap}, with less sensitivity to texture and color.
To address these challenges, we propose CoNav, a collaborative cross-modal reasoning framework for embodied navigation. In CoNav, the pretrained 3D–text model explicitly guides the image–text navigation agent by sharing textual spatial-semantic knowledge. To resolve conflicting beliefs among modalities, we perform lightweight fine-tuning of CoNav using only a small 2D–3D–text corpus, jointly performing the embodied navigation task. Specifically, (1) Separate Pretraining. We pretrain the image-text navigation agent on navigation image–text corpora and the 3D–text model on 3D–text pairs separately. (2) Cross‐Modal Belief Alignment fine-tuning. To allows the 3D–text model guide the navigation agent using text, we design a Communication Interface that enable information sharing. On this basis, we perform lightweight fine-tuning on a small 2D–3D–text corpus. The agent learns to reason effectively by fusing visual cues with spatial-semantic knowledge derived from the provided 3D information, resolving ambiguities during navigation.

In summary, the main contributions of this work are:
\begin{itemize}[leftmargin=8mm]
\item We introduce \textbf{CoNav}, a novel collaborative framework for embodied navigation that allows 3D–text model guide the navigation agent, by simply sharing textual hypotheses without requiring large-scale triple-modality datasets.

\item We demonstrate that Cross‐Modal Belief Alignment is an efficient way for embodied agents to integrate visual cues with spatial-semantic knowledge. Model-level fusion for 2D images and 3D point clouds can reduce the inappropriate actions due to the ambiguities relying only on pixel-level color and visual cues.

\item Extensive experiments demonstrate CoNav’s superior performance not only on embodied navigation benchmarks, but also on spatial reasoning tasks. This integration of multiple available modalities and collaborate to complete task is essential for building a practical and general embodied navigation agent.

\end{itemize}

\section{Related work}
\label{sec:rela}
\textbf{Vision-language navigation} (VLN). VLN aims to let the navigation agents follow linguistic navigation instructions and complete various navigation-related tasks \cite{anderson2018vision,ku2020room,zhu2021soon}. Initial VLN methods mainly focused on encoding vision and text, and their cross-modal connection. PRESS \cite{li2019robust} adopts BERT encoding instructions. NavGPT2 \cite{zhou2025navgpt} employs Q-former to link vision and text, with topological maps supporting action planning. 
Recently, LLMs is introduced to embodied navigation due to their strong language understanding. NavCoT \cite{lin2024navcot} fine-tunes LlaMA-7B with Chain of Thought to handle navigation tasks on text-only LLMs. NaviLLM \cite{zheng2023towards} fine-tunes Vicuna-7B \cite{vicuna2023} with multi-task learning. RoomTour3D \cite{han2024roomtour3d} enhances geometry-awareness by reconstructed 3D scenes. SAME \cite{zhou2024same} propose a new State-Adaptive Mixture of Experts model that enables an agent to infer decisions based on dynamic observations. SRDF \cite{wang2024bootstrapping} generates super large-scale navigational instruction-trajectory pairs, surpassing human performance in VLN). There are some graph base methods. GOAT \cite{wang2024vision} designed a causal transformer based on causal graph. SUSA \cite{zhang2024agent} combines depth and RGB Exploration Map. Despite their contributions, these methods rely on 2D visual cues. BSG \cite{liu2023bird} applies the Bird’s-Eye-View (BEV) concept, learned from multi-camera images in autonomous driving systems \cite{li2022bevformer}, to assist navigation. VER \cite{liu2024volumetric} generates structured 3D volumetric representations from multi-view images to help agents predict 3D occupancy \cite{song2017semantic}. 

\textbf{Injecting 3D into LLMs.}
Recent efforts have augmented language models with 3D geometric cues—ranging from point clouds to voxel grids and bird’s‐eye views. PointCLIP \cite{PointCLIPV2} and CLIP2Point \cite{CLIP2Point} use depth image projections of point clouds and apply 2D CLIP \cite{CLIP} models for 3D understanding. ULIP-2 \cite{ULIP-2} and OpenShape \cite{OpenShape} train point cloud encoders directly with point clouds and expand the training data through automatic generation via image captioning models. Cap3D \cite{Cap3D} generated 660K 3D-text pairs by applying the image captioning model to 3D objects in Objaverse \cite{Objaverse}, providing large-scale data for training 3D-LLMs with point clouds. PointLLM \cite{xu2025pointllm} improves the quality of 3D-text pairs and introduces new indicators to evaluate model performance. MSQA \cite{linghu2024multi} highlights the limitations of relying solely on vision cues. Many existing works such as \cite{xu2025pointllm, Point-BERT} focus on a single object. Sceneverse \cite{jia2024sceneverse} introduce large-scale indoor 3D scenes. Existing 3D indoor datasets often assume full access to 3D environments during inference. By contrast in realistic embodied navigation, agents could only incrementally and partially perceives 3D environment on fly, without prior full-scene observation.

\textbf{Collaborative Reasoning.} A growing body of work investigates multi-model collaboration, and these methods achieve model-level fusion. CoELA \cite{zhang2023building} integrates multiple LLMs as embodied agents that iteratively plan, communicate, and cooperate to complete navigation and manipulation tasks. Similarly, Embodied-R \cite{zhao2025embodied} pairs a large-scale vision–language model with a lightweight language model and, using only 5k training examples, matches the performance of state-of-the-art multimodal reasoning systems (e.g., OpenAI-o1, Gemini-2.5-pro). M500 \cite{jin2025two} use an extra "CEO" model to align multiple models and resolve conflicting beliefs between them. It fine-tune Qwen2.5-32B-Instruct on a collaborative reasoning traces dataset and let a CEO agent dynamically manages the discussion and collaboration.
\begin{figure*}[t]
    \centering
    \includegraphics[width=1\linewidth]{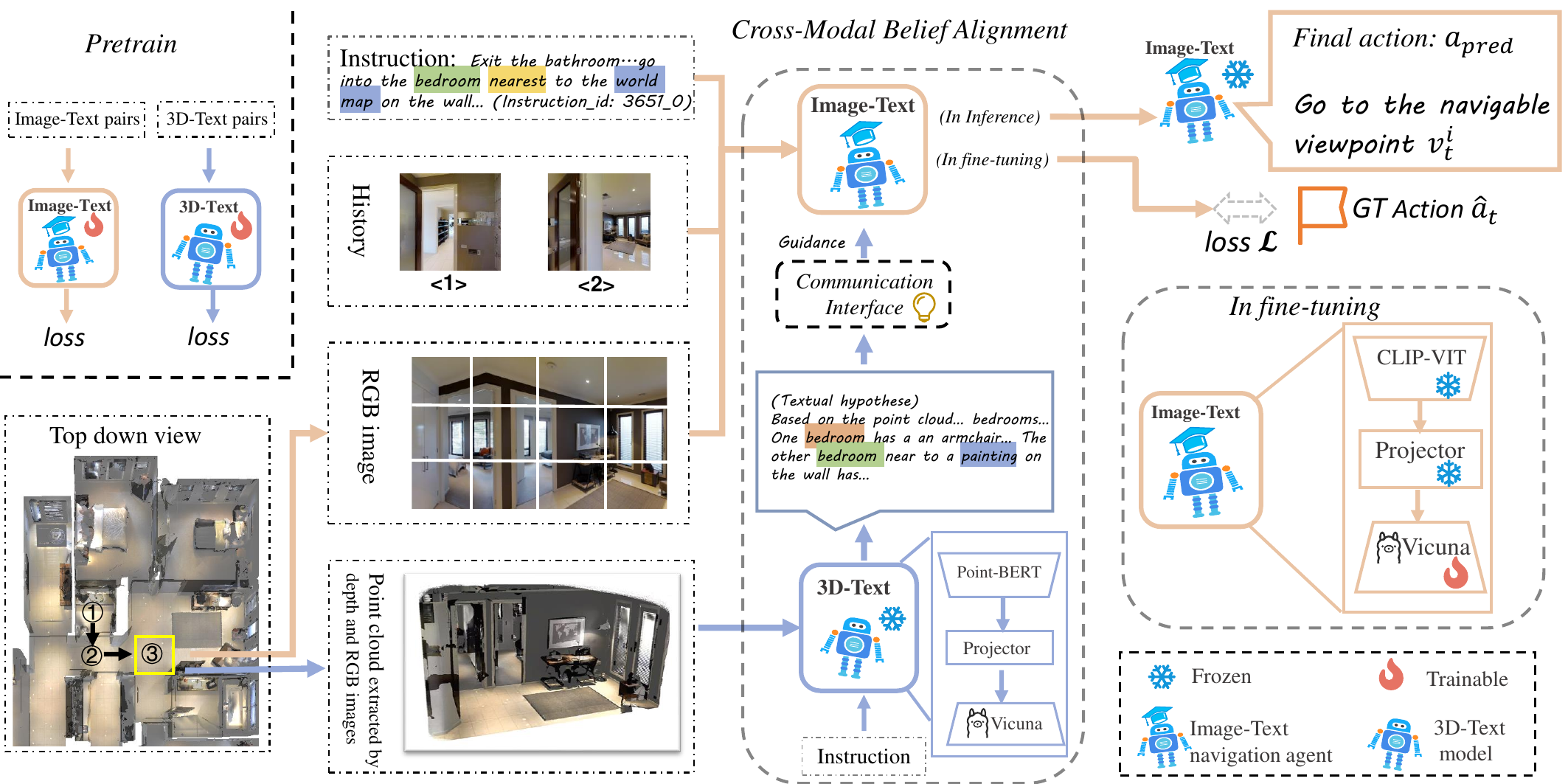} 
    \caption{ \small Our CoNav collaborative framework comprises an image–text navigation agent and a 3D–text model. The pretrained 3D–text model explicitly guides the image–text agent by providing structured spatial-semantic knowledge to resolve navigation ambiguities. The core of CoNav is the Cross-Modal Belief Alignment, which employs a Communication Interface to facilitate the straightforward sharing of textual hypotheses from our 3D–text model with our navigation agent. Through lightweight fine-tuning on a small 2D–3D–text corpus, CoNav learns to integrate visual cues with spatial-semantic knowledge derived from our 3D–text model.}
    \label{fig:pipline}
\end{figure*}
\section{Methods}
\label{sec:Approach}
In this section, we first introduce the navigation problem in (\S3.1). Then in (\S3.2), we describe how the image-text navigation agent and the 3D-text model are pretrained. After that, we present our \textit{Cross-Modal Belief Alignment} fine-tuning and inference in (\S3.3).

As previously mentioned, training a unified fusion model that integrates 2D, 3D, and text modalities presents several challenges. Studies \cite{zhang2023meta, han2025multimodal} and our preliminary exploration suggest that training a single unified fusion model is challenging in engineering practice. Therefore, we decouple the image–text navigation agent and the 3D–text model. This also enhances both flexibility and generalizability. Point clouds are more intuitive compared to voxels, BEV, and multiview, as they better represent the actual spatial structure of objects. Therefore, we use point clouds as the data for the 3D modality. Our framework is illustrated in Figure \ref{fig:pipline}.
\subsection{Problem Setup}
Given an instruction $L$, the embodied agent needs to navigate in a 3D environment according to the instruction. The environment consists of multiple 3D scenes $S$, $S = \{s_0,s_1, s_2, \dots, s_n\}$, where the agent starts at the scene $s_0$ and moves toward target scene $s_n$. At time $t$, the agent occupies scene $s_t$.  It observes an RGB–D frame $o_t = (r_t, d_t)$, where $r_t \in \mathbb{R}^{H\times W \times 3}, d_t \in \mathbb{R}^{H\times W}$ are the RGB and depth images from agent's camera.
From $o_t$, we extract a point cloud of the scene $p_t \in \mathbb{R}^{M \times 6}$ (3 for position and 3 for color), \(M\) is the number of points.
In VLN task, each state $s_t$ offers $K_t$ navigable viewpoints \(\mathcal{V}_t = \{\,v_t^1, v_t^2, \dots, v_t^{K_t}\},\) and the agent’s action space at $s_t$ is defined as \(\mathcal{A}_t = \mathcal{V}_t \,\cup\, \{\mathtt{stop}\},\) allowing the agent to move to any adjacent viewpoint or to terminate navigation. We denote the agent’s selection policy by 
\(
a_t = \pi\bigl(o_t,\,h_t,\,L\bigr), a_t\in\mathcal{A}_t,
\)
where \(h_t\) is the history of past observations and actions. Then the agent transitions to the next scene
\(s_{t+1}.\) Navigation is completed when the agent issues \(\mathtt{stop}\) or a maximum step count \(T_{maxStep}\) is reached. Agent also aim to minimize the route length \(T_{length}\) as possible to satisfy efficiency.

\subsection{Pretrain of image-text agent \& 3D-text model}
First of all, we need to prepare the pretrained navigation agent and 3D-text model. Both the navigation agent and the 3D-text model employ architectures based on widely-used multimodal large language model, often referred to as 'ViT-MLP-LLM' in existing studies \cite{zhu2023minigpt, LLaVA, xu2025pointllm}. In 'ViT-MLP-LLM', 'ViT' encoder (Enc) extracts features from either 2D images or 3D point clouds. MLP projector (Proj) maps these features into the text embedding space.
 
\textbf{Pretrain of navigation agent.} Initially, we pretrain the image–text navigation agent on diverse embodied navigation image–text pairs, following approaches in \cite{zheng2023towards, zhou2025navgpt}. We optimize the agent’s policy parameters \(\theta_{\text{nav}}\) by minimizing the following objective:
\begin{equation}
\begin{aligned}
  \pi_{\mathrm{nav}}(r_t, h_t, L)
  = \mathrm{LLM}^{\text{\TFicon{T}}}\Bigl(
       \mathrm{Proj}^{\text{\TFicon{T}}}\bigl(\mathrm{Enc}^{\text{\TFicon{F}}}(r_t,
       \,h_t)\bigr),\,L
     \Bigr),\\
  \min_{\theta_{\mathrm{nav}}}\ 
  \mathbb{E}_{(r_t, h_t, L, \hat a_t)\sim\mathcal{D}_{2D}}
    \Bigl[\,
      \mathcal{L}\bigl(\pi_{\mathrm{nav}}(r_t, h_t, L),\,\hat a_t\bigr)
    \Bigr].
\end{aligned}
\end{equation}

Where \(\theta_{\text{nav}}\) represents the trainable parameters of the navigation agent. \TFicon{F} indicates a frozen component, and \TFicon{T} indicates a trainable component. \(r_t\) is the RGB image at time step \(t\), \(h_t\) is the history of past observations and actions, and \(L\) is the instruction text. \(\hat{a_t}\) is the ground truth (GT) action at time step \(t\). \(\mathcal{D}_\text{2D}\) denotes the image–text pretraining dataset. \(\pi_{\text{nav}}(r_t, h_t, L)\) represents the navigation agent’s policy, and its output belongs to the action space \(\mathcal{A}_t\). \(\mathcal{L}(\pi_{\text{nav}}(r_t, h_t, L), \hat{a_t})\) is the loss function measures the difference predicted action \(\pi_{\text{nav}}(r_t, h_t, L)\) and the actual action \(\hat{a_t}\). 

\textbf{Pretrain of 3D-text model.}
The 3D-text model also follows the ‘ViT-MLP-LLM’ structure.
Since spatial reasoning in real scanned point cloud scenes is challenging especially in embodied navigation, we designed a Progressive Curriculum Learning \cite{bengio2009curriculum} pretrain paradigm to overcome these challenges (See Appendix \ref{sec:append_3dllm} for more details). Curriculum Learning \cite{bengio2009curriculum} refers to imitating the characteristics of human learning and learning from simple to difficult. This makes it easier for the model to find a better local optimum and speeds up training and it is widely used in such LLaVA-OneVision\cite{li2024llava} and Lavie \cite{wang2025lavie}.
We divide the pretraining into \textit{three distinct stages} using different datasets from simple to complex. The overall pretrain process of 3D-text model is as follows:
\begin{equation}\label{equ:align_opt}
\begin{gathered}
  \pi_{3D}(p, L)
  = \mathrm{LLM}\Bigl(
      \mathrm{Proj}\bigl(\mathrm{Enc}(p)\bigr),\,L
    \Bigr),\\[1ex]
  \min_{\theta_{3D}}
  \mathbb{E}_{(p, L, \hat a_t)\sim\mathcal{D}^{i}_{3D}}
    \Bigl[
      \mathcal{L}\bigl(\pi_{3D}(p,L),\,\hat a_t\bigr)
    \Bigr],\quad
  \mathcal{D}^{i}_{3D}\in
  \bigl\{\mathcal{D}^{stage1}_{3D},\,\mathcal{D}^{stage2}_{3D},\,\mathcal{D}^{stage3}_{3D}\bigr\}.
\end{gathered}
\end{equation}
\(\pi_{\text{3D}}\) represents 3D-text model’s action. The term \(\mathcal{D}^{i}_\text{3D}\) refers to the pretraining dataset in current stage. 
\subsection{Cross-Modal Belief Alignment} 
In Cross-Modal Belief Alignment fine-tuning, we enable cross-modal guidance by simply sharing textual hypotheses from the 3D–text model with the navigation agent. To support this, we design a Communication Interface that organize hypotheses into the standard prompt format as shown:
\begin{equation}\label{equ:prompt}
\makebox[\textwidth][c]{%
  \begin{minipage}[t]{0.45\textwidth}
    \centering
    {\footnotesize Standard prompt format in image–text agent}\\[1ex]
    \resizebox{\textwidth}{!}{$
      \begin{aligned}
        &\underbrace{\text{\texttt{You are a helpful...}}}_{\text{System message}},\,
         \underbrace{\text{\texttt{Your task is...}}}_{\text{Task instruction}},\\
        &\underbrace{r_i^1, r_i^2, \dots}_{\text{2D RGB tokens}},\,
         \underbrace{\pi_{\text{3D}}}_{\substack{\text{Text hypothesis}\\\text{from 3D-text model}}},\,
         \underbrace{\text{\texttt{Select the...}}}_{\text{Output hint}}.
      \end{aligned}
    $}
  \end{minipage}%
  \hfill
  \begin{minipage}[t]{0.45\textwidth}
    \centering
    {\footnotesize Standard prompt format in 3D–text model}\\[1ex]
    \resizebox{\textwidth}{!}{$
      \begin{aligned}
        &\underbrace{\text{\texttt{A chat between...}}}_{\text{System message}},\,
         \underbrace{\text{\texttt{Your task is...}}}_{\text{Task instruction}},\\
        &\underbrace{p_i^1, p_i^2, \dots}_{\text{3D point cloud tokens}},\,
         \underbrace{\text{\texttt{Illustrate the spatial...}}}_{\text{Output hint}}.
      \end{aligned}
    $}
  \end{minipage}%
}
\end{equation}

For brevity, Eq. \ref{equ:prompt} only shows an overview of the Standard prompt format in image-text navigation agent and 3D-text model. You can find the detailed process of the prompt format and Communication Interface organize hypotheses into the standard prompt format in the Appendix \ref{sec:appendix_prompt}. The pretrained 3D-text model in Cross-Modal Belief Alignment describes the spatial-semantic structure of the scene when tackling an embodied navigation task, and answers spatial reasoning questions when addressing spatial reasoning tasks. The navigation agent receives the textual hypothesis \(\pi_{\text{3D}}\) from 3D-text model and then the navigation agent generates the final action. The Communication Interface enables cross-modal guidance from the 3D–text model to the navigation agent, achieving a model-level fusion for CoNav.

To minimize the need for large triple‐modality datasets and reduce training costs, we freeze all parameters of the 3D–text model (\(\theta_{\text{3D}}\)) and fine‐tune only a lightweight subset of the navigation agent’s LLM parameters (\(\theta_{\mathrm{nav}}\)) on a small triple‐modality corpus \(\mathcal{D}_{\mathrm{3MT}}\). Thus, Belief Alignment fine‐tuning seeks to preserve each model’s unique prior knowledge through minimal updates, while simultaneously resolving conflicting beliefs between the navigation agent and the 3D–text model.

During Belief Alignment fine-tuning at each time step \(t\), the navigation agent produces a task instruction for the 3D–text model based on the current task. The 3D–text model then generates the action \(\pi_{\text{3D}}\), a spatial hypothesis in natural-language based on point cloud \(p_t\) and instruction text \(L\). The navigation agent receives \(\pi_{\text{3D}}\) through the text‐level Communication Interface and generates its action $a_t$. Finally, we lightweight fine-tune the trainable LLM parameters \(\theta_{\mathrm{nav}}\) in navigation agent by minimizing the following supervised loss over the small triple‐modality dataset \(\mathcal{D}_{\mathrm{3MT}}\):
\begin{equation}\label{equ:align_opt_updated}
\begin{gathered}
  \pi_{\mathrm{nav}}\bigl(r_t,h_t,\mathrm{Fmt}(\pi_{\text{3D}}(p_t,L),L)\bigr)
  = \mathrm{LLM}^{\text{\TFicon{T}}}\Bigl(
       \mathrm{Proj}^{\text{\TFicon{F}}}\bigl(\mathrm{Enc}^{\text{\TFicon{F}}}(r_t,\,h_t)\bigr),\,
       \mathrm{Fmt}\bigl(\pi_{\text{3D}}(p_t,L),L\bigr)
     \Bigr),\\[1ex]
  \min_{\theta_{\mathrm{nav}}}
  \;\mathbb{E}_{(r_t,p_t,h_t,L,\hat a_t)\sim\mathcal{D}_{\mathrm{3MT}}}
    \Bigl[
      \mathcal{L}\bigl(\pi_{\mathrm{nav}}(r_t,h_t, \mathrm{Fmt}(\pi_{\text{3D}}(p_t,L),L)),\,\hat a_t\bigr)
    \Bigr].
\end{gathered}
\end{equation}

Here, \(\hat{a}_t\) is the ground‐truth action at time \(t\), and \(\mathcal{L}\) is a is the loss function optimize the predicted and actual actions. \(\mathrm{Fmt}(\cdot)\) denotes the standard formatting operator (Eq.~\ref{equ:prompt}) by Communication Interface. By freezing \(\theta_{\text{3D}}\) and only fine-tuning \(\theta_{\mathrm{nav}}\), CoNav aligns cross‐modal beliefs using minimal trainable parameters and triple‐modality data. 
 
 At inference time step \(t\), agent observes the instruction text \(L\), the image \(r_t\) and the point cloud \(p_t\) in the scene. The 3D model processes \((p_t, L)\). The result spatial hypothesis in natural-language \(\pi_{\text{3D}}\) is sent to the navigation agent. Then the navigation agent \(\pi_{\mathrm{nav}}\) processes \(r_t, L\) and \(\pi_{\text{3D}}\), generating the final action \(a_{pred}\). The joint reasoning is computed as:
\begin{equation}\label{equ:inf}
a_{pred} \;=\;\pi_{\mathrm{CoNav}}\Bigl(
 r_t,\;h_t,\;\mathrm{Fmt}\bigl(\pi_{\text{3D}}(p_t,L),\,L\bigr)
\Bigr),\quad
a_{pred} \in \mathcal{A}_t.
\end{equation}
Here, $\mathcal{A}_t$ represents the agent's action space at step $t$. The policy $\pi_{\mathrm{CoNav}}$ fuses the navigation agent with the spatial-semantic knowledge guidance for 3D-text model at model-level. The 3D-text model share textual hypotheses to navigation agent and jointly perform the embodied navigation task until the navigation agent issues \(\mathtt{stop}\) or the maximum step count \(T_{maxStep}\) is reached.

\section{Experiments}
\label{sec:experiments}
\subsection{Benchmarks and Metrics}  
\textbf{For embodied navigation.}
\textbf{i) Dataset}. We follow the previous NaviLLM \cite{zheng2023towards}, using four datasets to evaluate the performance of our CoNav on the embodied navigation task: R2R \cite{anderson2018r2r} provides detailed, step-by-step instructions. CVDN \cite{thomason2020vision} frames navigation as a dialog, requiring the agent to interpret and act on the dialog. REVERIE \cite{qi2020reverie} provides concise, high-level goal descriptions, forcing the agent to infer intermediate path. SOON \cite{zhu2021soon} focuses on object-centric navigation, where the agent locate the target based description of the target.
\textbf{ii) Metrics.} We adopt a comprehensive metrics, including Trajectory Length (TL) measures average distance traveled in meters, Navigation Error (NE) measures average distance between the final and target locations, Success Rate (SR) measures percentage of paths where the distance to final destination is within a predefined distance threshold, Success Rate Weighted by Path Length (SPL) measures SR adjusted by the ratio of the agent’s executed path length to the shortest ground-truth path length, reflecting whether the path is a "shorter path" and balancing SR with efficiency:
\begin{equation}
\mathrm{SPL} = \frac{1}{N} \sum_{i=1}^N S_i \,\frac{T_i}{T_{\text{pred}}}, \quad S_i = \begin{cases} 
1, & \text{if success}, \\
0, & \text{if failure}.
\end{cases} \quad S_i \in \{0, 1\}, \quad T_{\text{pred}} \geq T_i
\end{equation}
\(T_i\) is the ground-truth path length, N is the number of episodes and \(T_{\text{pred}}\) is the length of the agent’s executed path. \(\mathrm{SPL}\) takes into account both the success rate and the efficiency of the path. Goal Process (GP) measures the distance towards the target. 
 \textbf{iii) Baselines.} We compare our method with the \textit{open source} SOTA navigation method on R2R, CVDN, REVERIE and SOON. We do not consider methods with pre-exploration (e.g., AuxRN [60], RREx-BoT [50]) and task-specialized method (e.g., GOAT \cite{wang2024vision}, SUSA \cite{zhang2024agent}) requiring separate training on different datasets. 

\textbf{For spatial reasoning.}
\textbf{i) Dataset}. We utilize the ScanQA \cite{azuma2022scanqa} and SQA3D \cite{ma2022sqa3d} datasets to evaluate spatial reasoning and understanding abilities. ScanQA \cite{azuma2022scanqa} focuses on object-grounded question answering in 3D indoor scenes, while SQA3D \cite{ma2022sqa3d} involves situated question answering, requiring agents to comprehend their position and orientation within a 3D scene.
\textbf{ii) Metrics.} We use Exact Match, METEOR, BLEU and ROUGE-L to evaluate performance on ScanQA and Exact Match for SQA3D following previous method \cite{huang2023embodied}.
\textbf{iii) Baselines.} We compare with recently proposed \textit{open source} 3D-LLMs such as SceneVerse \cite{jia2025sceneverse}, LEO \cite{huang2023embodied} and 3D-VisTA \cite{zhu20233d}. 

\subsection{Implementation Details} 
In navigation task, we divide the data into val-unseen (Val-U) split and test split. The Val-U split measures the performance locally, for fair the test split measures the performance on an AI model evaluation online platform\footnote{\tiny https://eval.ai}, follow the previous approaches. You can find our test split results public on the platform\footnote{\tiny R2R: https://eval.ai/web/challenges/challenge-page/97/overview, ScanQA: https://eval.ai/web/challenges/challenge-page/1715/overview, \\ REVERIE: https://eval.ai/web/challenges/challenge-page/606/overview, CVDN: https://eval.ai/web/challenges/challenge-page/463/overview.} on the leaderboard.

\textbf{In the navigation model}, we use ViT-g/14 from EVA-CLIP \cite{fang2023eva} as the image encoder. We follow NaviLLM \cite{zheng2023towards} to pretrain the navigation agent. Images and texts are unified into 4096-dimensional tokens, which are then input into the LLM. We used Vicuna-7B \cite{vicuna2023} for both navigation agent and 3D-text model. \textbf{In 3D-text model}, we use pretrained Point-BERT \cite{Point-BERT} as our point encoder. For each point cloud, we uniformly sample 8192 points and group the point cloud into 512 patches. Point cloud and text are unified into 4096-dimensional tokens. In \textbf{Cross-Modal Belief Alignment} we use triples from R2R, CVDN, REVERIE, SOON, ScanQA and SQA with a sampling ratio of 4:1:2:1:1:1. You can find detailed training times hyperparameters and data visualization in Appendix \ref{sec:append_hyperparameters}, \ref{sec:append_Task_example}.
\begin{table*}[tb]
    \centering
    \caption{\small Overall quantitative comparison of SOTA methods on both embodied navigation and spatial reasoning tasks. For embodied navigation, we report SPL for R2R, REVERIE and SOON, GP for CVDN. For spatial reasoning, we report Exact Match for ScanQA and SQA3D, on Val-U and test split. CoNav achieves state-of-the-art performance across different navigation tasks. While the GP improvement on the CVDN Test set is modest, our CoNav outperforms NaviLLM \cite{zheng2023towards} by approximately 56\% (0.09 -> 0.14) in SPL on the Test set, reflecting CoNav can find a shorter path. * denotes reproduced results.}
\resizebox{0.99\textwidth}{!}{%
\begin{tabular}{c|cc|cc|cc|cc|cc|c} 
    \toprule
\multirow{2}{*}{\textbf{Method}} 
  & \multicolumn{2}{c|}{\textbf{R2R}} 
  & \multicolumn{2}{c|}{\textbf{CVDN}} 
  & \multicolumn{2}{c|}{\textbf{REVERIE}} 
  & \multicolumn{2}{c|}{\textbf{SOON}}
  & \multicolumn{2}{c|}{\textbf{ScanQA}} 
  & \multicolumn{1}{c}{\textbf{SQA3D}}\\
  & $\text{Val-U}^{\uparrow}$ 
  & $\text{Test}^{\uparrow}$ 
  & $\text{Val-U}^{\uparrow}$ 
  & $\text{Test}^{\uparrow}$ 
  & $\text{Val-U}^{\uparrow}$ 
  & $\text{Test}^{\uparrow}$ 
  & $\text{Val-U}^{\uparrow}$ 
  & $\text{Test}^{\uparrow}$ 
  & $\text{Val-U}^{\uparrow}$ 
  & $\text{Test}^{\uparrow}$ 
  & $\text{Test}^{\uparrow}$\\ 
    \hline
    \multicolumn{12}{c}{\textit{\textbf{Embodied Navigation Model}}}\\ 
    \hline
PREVALENT \cite{hao2020towards}
  & 53 & 51 & 3.15 & 2.44 & - & - & - & - & - & - & -\\
HOP \cite{qiao2022hop}                         
  & 57 & 59 & 4.41 & 3.24 & 26.1 & 24.3 & - & - & - & - & -\\ 
HAMT \cite{NEURIPS2021_2e5c2cb8}                        
  & 61 & 60 & 5.13 & 5.58 & 30.2 & 26.7 & - & - & - & - & -\\ 
DUET \cite{chen2022duet}                        
  & 60 & 58 & -    & -    & 33.7 & 36.1 & 22.6 & 21.4 & - & - & -\\ 
VLN-SIG \cite{li2023improving}
  & 62 & 60 & 5.52 & 5.83 & - & - & - & - & - & - & -\\
VLN-PETL \cite{qiao2023vln}                        
  & 60 & 58 & 5.69 & 6.13 & 27.7 & 26.7 & - & - & - & - & -\\ 
BEV-BERT \cite{an2022bevbert}                        
  & 64 & 60 & -    & -    & \underline{36.4} & \textbf{36.4} & - & - & - & - & -\\ 
NavGPT-2 \cite{zhou2025navgpt}                        
  & \underline{61} & 60 & -    & -    & - & - & - & - & - & - & -\\ 
NaviLLM \cite{zheng2023towards}                        
  & 59 & \underline{60} & \underline{6.16} & \underline{7.90} 
  & 35.7 & 32.3 & \underline{29.2} & \underline{26.3} 
  & \underline{23.0} & \underline{24.8} & -\\ 
NaviLLM$^{*}$      
  & 57 & - & 6.09 & - & 31.4 & - & 28.0 & - & 22.4 & - & -\\
    \hline
\multicolumn{12}{c}{\textit{\textbf{3D Large Language Model}}}\\ 
    \hline
3D-LLM \cite{3D-LLM}                        
  & - & - & - & - & - & - & - & - & 20.5 & 19.1 & -\\ 
SceneVerse \cite{jia2025sceneverse}                        
  & - & - & - & - & - & - & - & - & 22.7 & 23.5 & \textbf{49.9}\\ 
    \hline
\rowcolor{Gray!20}
CoNav                      
  & \textbf{62} & \textbf{62} & \textbf{7.09} & \textbf{7.98} 
  & \textbf{37.5} & \underline{36.2} 
  & \textbf{30.0} & \textbf{27.2} 
  & \textbf{23.2} & \textbf{25.1} & \underline{49.0}\\  
    \bottomrule
\end{tabular}
}
\label{tab:methods_comparison}
    \vspace{-0.2cm}
\end{table*}

\subsection{Overall comparison} 
\label{sec:sota_cmp}  
\textbf{Overall comparison. }The overall result is shown in Table \ref{tab:methods_comparison}. CoNav is not only capable of performing embodied navigation tasks but also demonstrates strong proficiency in spatial reasoning. From the Table, navigation performance is related to spatial reasoning performance to some extent. Notably, our CoNav achieves state-of-the-art performance on both the Val-U and Test sets across those standard embodied navigation benchmarks. While the GP improvement on the CVDN Test set is modest compare to top 2, our CoNav outperforms by approximately 56\% (0.09 -> 0.14) in SPL on the Test set on the CVDN Leaderboard (Table \ref{table:cvdn_sota}), reflecting that the path generated by CoNav are "shorter paths" compared to other. We hypothesize that the improvement in navigation performance is attributed to the enhanced spatial reasoning capabilities. When visual cues lead to confusion, the navigation agent is guided by spatial-semantic knowledge derived from 3D-text model, avoiding the inappropriate actions of relying on pixel-level color and visual cues. 

\textbf{Path Efficiency and Performance Insights.} 
As shown in Table \ref{table:r2r_sota}, for R2R step-by-step instructions, although CoNav performs slightly worse than NavGPT-2 (FlanT5-11B)\cite{zhou2025navgpt} in terms of SR, it achieves a higher SPL. This indicates that CoNav tends to generate shorter paths. Furthermore, when both navigation agents use Vicuna-7B, our method shows a significant improvement of approximately 20\% over NavGPT-2, suggesting that the 3D-text model in CoNav provides better guidance for Vicuna-7B in navigation agents, thereby enhancing the navigation performance. 
Similar conclusions can be drawn from Table \ref{table:reverie_sota} and Table \ref{table:cvdn_sota}, where CoNav also favors shorter paths. In the CVDN test split from Table \ref{table:cvdn_sota}, despite our GP being slightly lower than the top 1 SRVLN on the leaderboard, our SPL is even more than twice that of SRVLN. Increasing the maximum step count \(T_{maxStep}\) in navigation can improve SR to some extent, but it does not lead to shorter paths and reduces SPL. A good navigation agent needs not only to successfully reach the destination but also to take the shortest path. 

We hypothesize that this improvement is due to the prior knowledge provided by the 3D-text model, which helps the navigation agent plan more reasonable paths based on the volumetric structure of the scene, resulting in shorter paths to the destination compared to other methods. For example, when multiple navigable paths to the target are available, the agent can select the shortest one guided by spatial distance reasoning from 3D-text model.
\begin{table*}[tbp!]
\begin{minipage}{0.5\linewidth}
\centering
\captionsetup{width=0.95\textwidth}
\caption{\small Detailed comparison with SOTA methods on the R2R dataset. Specialists means these methods are initialized from general vision-language models, and incorporate VLN-tailored pretraining.}
    \vspace{-0.2cm}
\resizebox{\linewidth}{!}{%
\begin{tabular}{l|cccc|cccc} 
\toprule
\multirow{2}{*}{\textbf{Method}}
  & \multicolumn{4}{c|}{Val Unseen}
  & \multicolumn{4}{c}{Test Unseen} \\
  & TL
  & $\text{NE}^{\downarrow}$
  & $\text{SR}^{\uparrow}$
  & $\text{SPL}^{\uparrow}$
  & TL
  & $\text{NE}^{\downarrow}$
  & $\text{SR}^{\uparrow}$
  & $\text{SPL}^{\uparrow}$ \\ 
\hline
Human\cite{zhou2025navgpt} & -    & -    & -    & -    & 11.85 & 1.61 & 86 & 76 \\
Seq2Seq\cite{anderson2018r2r} & 8.39 & 7.81 & 21   & -    & 8.13  & 7.85 & 20 & -  \\
RCM\cite{wang2019reinforced}   & 11.46 & 6.09 & 43   & -    & 11.97 & 6.12 & 43 & 38 \\
EnvDrop\cite{tan2019envdrop}   & 10.70 & 5.22 & 52   & 48   & 11.66 & 5.23 & 51 & 47 \\ \hline
\multicolumn{9}{c}{\textit{\textbf{Specialists with Vision-Language-Action Pretraining}}} \\ \hline
PREVALENT \cite{hao2020towards} & 10.19 & 4.71 & 58   & 53   & 10.51 & 5.30 & 54 & 51 \\
HOP \cite{qiao2022hop}          & 12.27 & 3.80 & 64   & 57   & 12.68 & 3.83 & 64 & 59 \\
VLN-BERT \cite{Hong_2021_CVPR}  & 12.01 & 3.93 & 63   & 57   & 12.35 & 4.09 & 63 & 57 \\
DUET \cite{chen2022duet}        & 13.94 & 3.31 & 72   & 60   & 14.73 & 3.65 & 69 & 59 \\
Meta-Explore \cite{Hwang_2023_CVPR}&13.09&3.22&72   & 62   & 14.25 & 3.57 & 71 & 61 \\
AZHP \cite{Gao_2023_CVPR}       & 14.05 & 3.15 & 72   & 61   & 14.95 & 3.52 & 71 & 60 \\
VLN-PETL \cite{qiao2023vln}     & 11.52 & 3.53 & 65   & 60   & 12.30 & 4.10 & 63 & 58 \\ \hline
\multicolumn{9}{c}{\textit{\textbf{LLM-based navigation (zero-shot)}}}      \\ \hline
MapGPT\cite{chen2024mapgpt}     & -     & 6.92 & 39   & 26   & -     & -    & -  & -  \\
DiscussNav\cite{long2023discuss}& 9.69  & 5.32 & 43   & 40   & -     & -    & -  & -  \\ \hline
\multicolumn{9}{c}{\textit{\textbf{LLM-based navigation}}}                  \\ \hline
NaviLLM$^{*}$ \cite{zheng2023towards}        & 14.02 & 3.65 & 66 & 57 & -    & -    & - & - \\
Navillm \cite{zheng2023towards}          & 12.81 & 3.51 & 67 & 59 & 11.74& 3.68 & 68& 60\\
NavGPT-2 (Vicuna-7B)\cite{zhou2025navgpt} & 12.29 & 4.86 & 54 & 45 &   -  & -    & - & - \\
NavGPT-2 (FlanT5-11B)\cite{zhou2025navgpt}& 13.57 & \textbf{3.13} & \textbf{72} & 61 & 14.58 & \textbf{3.35} & \textbf{71} & 60 \\ \hline
\rowcolor{Gray!20}
CoNav (Vicuna-7B)                        & 11.92 & 3.42 & 69 & \textbf{62} & 13.21 & 3.71 & 68 & \textbf{62} \\
\bottomrule
\end{tabular}%
\label{table:r2r_sota}
}
\end{minipage}
\begin{minipage}{0.46\linewidth}
\begin{minipage}{\linewidth}
\centering
\caption{\small Quantitative comparison with methods on spatial reasoning. “M” stands for “METEOR”, “R” for “ROUGE”, “B-4” for “BLEU-4” and “E” for “Exact Match”.}
    \vspace{-0.2cm}
\resizebox{\linewidth}{!}{
\begin{tabular}{l|cccccccc|c} 
    \toprule
\multirow{2}{*}{\textbf{Method}}
  & \multicolumn{4}{c}{\textbf{ScanQA Val-U}}
  & \multicolumn{4}{c|}{\textbf{ScanQA test}}
  & \multicolumn{1}{c}{\textbf{SQA3D}}\\ 
  & $\text{M}^{\uparrow}$ & $\text{R}^{\uparrow}$ & $\text{B-4}^{\uparrow}$ & $\text{E}^{\uparrow}$
  & $\text{M}^{\uparrow}$ & $\text{R}^{\uparrow}$ & $\text{B-4}^{\uparrow}$ & $\text{E}^{\uparrow}$
  & $\text{E}^{\uparrow}$\\ \hline
LLaVA \cite{LLaVA}                           & 10.5  & 12.3  & 0.3  & 0   & - & - & -  & -  & -   \\ 
Flamingo \cite{alayrac2022flamingo}          & 11.3  & 31.1  & 8.4  & 18.8& - & - & -  & -  & -   \\ 
VoteNet+MCAN                               & 11.4  & 29.8  & 6.2  & 17.3& 12.0& 30.9& 6.0& 19.7& 36.2\\ 
ScanQA \cite{azuma2022scanqa}               & 13.1  & 33.3  &10.1 & 21.1& 13.5& 34.3&12.0& 23.5& 47.2\\ 
3D-LLM \cite{3D-LLM}                        & 14.5  & 35.7  &12.0 & 20.5& 14.9& 35.3&11.6& 19.1& -   \\ 
3D-VisTA \cite{zhu20233d}                   & 13.9  & 35.7  &10.4 & 22.4& 12.9& 32.8&11.9& 23.0& 48.5\\ 
SceneVerse \cite{jia2025sceneverse}         & -     & -     & -   & 22.7& -   & -   & -  & 23.5& 49.9\\ \hline
LEO \cite{huang2023embodied}                & 20.0  & 49.2  &13.2 & 24.5& -   & -   & -  & -   &50.0\\ 
\rowcolor{Gray!20}
CoNav                                      & 15.2  & 38.9  &12.8 & 23.2& 15.2& 37.6&11.4& 25.1& 49.0\\       
\bottomrule
\end{tabular}
}
\label{tab:methods_comparison_3d_}
\end{minipage}
\begin{minipage}{\linewidth}
\caption{\small Detailed comparison with SOTA methods on REVERIE dataset.}
\centering
    \vspace{-0.2cm}
\resizebox{\linewidth}{!}{
\setlength{\tabcolsep}{3.3mm}{
\begin{tabular}{l|cccc|cccc} \toprule
\multirow{2}{*}{\textbf{Method}}
  & \multicolumn{4}{c|}{Val Unseen}
  & \multicolumn{4}{c}{Test Unseen} \\
  & \textbf{TL} 
  & \textbf{OSR} 
  & $\textbf{SR}^{\uparrow}$ 
  & $\textbf{SPL}^{\uparrow}$
  & \textbf{TL} 
  & \textbf{OSR} 
  & $\textbf{SR}^{\uparrow}$ 
  & $\textbf{SPL}^{\uparrow}$ \\ \midrule

Seq2Seq \cite{anderson2018vision} & 11.07 & 8.07 & 4.20 & 2.84 & 10.89 & 6.88 & 3.99 & 3.09 \\
HOP \cite{qiao2022hop}             & 18.85 & 36.24& 31.78& 26.11& 16.38 & 33.06& 30.17& 24.34 \\
HAMT \cite{NEURIPS2021_2e5c2cb8}    & 14.08 & 36.84& 32.95& 30.20& 13.62 & 33.41& 30.40& 26.67 \\
VLN-BERT \cite{Hong_2021_CVPR}      & 16.78 & 35.02& 30.67& 24.90& 15.68 & 32.91& 29.61& 23.99 \\
DUET \cite{chen2022duet}           & 22.11 & 51.07& 46.98& 33.73& 21.30 & 56.91& \textbf{52.51}& 36.06 \\
AZHP \cite{Gao_2023_CVPR}          & 22.32 & 53.65& \textbf{48.31}& 36.63& 21.84 & 55.31& 51.57& 35.85 \\
VLN-PETL \cite{qiao2023vln}        & 14.47 & 37.03& 31.81& 27.67& 14.00 & 36.06& 30.83& 26.73 \\ \hline
\rowcolor{Gray!20}
CoNav                               & 14.01 & 50.34& 42.96& \textbf{37.54}& 14.20 & 49.46& 43.04& \textbf{36.16}\\\hline
\end{tabular}
}}
\label{table:reverie_sota}
\end{minipage}
\end{minipage}
\end{table*}

\subsection{Ablation Study}
\textbf{Effectiveness of Model-level Fusion}. As shown in Table \ref{table:Ablation_study}, first, from settings i and ii, it is clear that relying solely on the point cloud–text model outperforms relying solely on the image–text model in spatial reasoning tasks. Further, as shown in setting iv, the fusion of all three modalities outperforms both in embodied navigation tasks and spatial reasoning. This may be because model-level fusion leverages each model’s distinct strength. Model-level fusion also enables implicit cross-verification: the navigation agent can use volumetric structure to cross-check visual cues and avoid potential ambiguities. Since relying solely on 3D-text cannot complete the navigation task, the result for R2R in setting ii is empty. This highlights the effectiveness of our model-level fusion in embodied navigation.
\begin{wraptable}{r}{0.3\linewidth}
\captionsetup{labelfont=small}
\caption{\footnotesize Top on CVDN Leaderboard Until May 2025.}
\resizebox{\linewidth}{!}{
\begin{tabular}{l|l|cccccc} \hline
Rank&Team&\textbf{SPL}$\uparrow$&\textbf{GP}$\uparrow$ \\ \toprule
1&SRVLN &0.06&8.19\\
\rowcolor{Gray!20}
2&CoNav	&\textbf{0.14}	&	7.98\\
3&NaviLLM&0.09&7.90\\
4&RoomTour3D&0.12&7.55\\
5&WCGen3&0.07&7.18\\
\bottomrule
\end{tabular}
\label{table:cvdn_sota}
}
\end{wraptable}
\textbf{Effectiveness of Cross-Modal Belief Alignment}. As shown in setting iii, simply sharing textual hypotheses from the 3D-text model with the navigation agent without alignment, leads to a performance drop in both ScanQA and R2R. This may be due to conflicting beliefs among modalities. Our Cross-Modal Belief Alignment helps resolve these conflicts, effectively integrating visual cues with spatial-semantic knowledge. Therefore, effective our Cross-Modal Reasoning relies on the Cross-Modal Belief Alignment, and both are essential for helping our CoNav understand the 3D world.

\subsection{More Analysis}
\textbf{Cross-Modal Alignment Analysis.} Some works, such as Embodied-R \cite{zhao2025embodied}, achieve alignment and collaborative reasoning between multiple agents at model-level using reinforcement learning. However, these methods require carefully designed reward functions, whereas CoNav does not. Additionally, methods like M500 \cite{jin2025two} use an extra "CEO" model to align multiple agents at model-level and resolve conflicting beliefs between them. We explore the use of a Qwen2.5-7B \cite{yang2024qwen2} as the CEO model to align our navigation agent with the 3D-text model without training. As shown in Table \ref{table:fusion_level}, the Post-Process Alignment approach use an extra "CEO" does not perform as well as our Cross-Modal Belief Alignment.

\textbf{Feature-level Fusion \textit{vs.} Model-level Fusion.} Cross-modal fusion can occur at different levels. As shown in Table~\ref{table:fusion_level}, here we simultaneously use the point encoder with its projector and the image encoder with its projector from CoNav. The projector outputs are fused and passed into the LLM, achieving feature-level fusion. We fine-tune this feature-level fusion model on the same small triple-modality dataset used for Cross-Modal Belief Alignment. Compared to model-level fusion, feature-level fusion performs worse. This may be because unified feature-level fusion requires more triple-modality data to fully realize its potential on embodied navigation task here. This result underscores the practical value of CoNav in settings with limited data availability.

\begin{table*}[btp!]
\begin{minipage}{0.65\linewidth}
\captionsetup{labelfont=small}
\centering
\captionsetup{width=0.95\textwidth}
\caption{\small Ablation study. "M" stands for "METEOR", "R" for "ROUGE". Modality represents the modality used. In setting iii, we simply share information from the 3D-text model to the navigation agent without Cross-Modal Belief Alignment fine-tuning. Effective cross-modal reasoning depends on successful Cross-Modal Belief Alignment fine-tuning.}
\resizebox{\textwidth}{!}{
\begin{tabular}{c|c|c|ccc|cc|cc}
    \toprule
 &\textbf{Cross-Modal} &  \textbf{ Model-level} & \multicolumn{3}{c|}{\textbf{Modality}} & \multicolumn{2}{c|}{\textbf{ScanQA}} & \multicolumn{2}{c}{\textbf{R2R}} \\
& \textbf{Belief Alignment}&\textbf{Fusion} &Text&Image&Point cloud&M$\uparrow$&R$\uparrow$&SPL$\uparrow$&SR$\uparrow$\\ \hline
i.& \ding{55}&\ding{55}&\ding{51}&\ding{51}&\ding{55}& 14.3  & 36.7&57&64 \\
ii.&\ding{55}&\ding{55}&\ding{51}&\ding{55}&\ding{51}& 14.5 & 37.1 & - & -\\
iii.&\ding{55}&\ding{51}&\ding{51}&\ding{51}&\ding{51}&14.6&35.7&56&63\\
iv.&\ding{51}&\ding{51}&\ding{51}&\ding{51}&\ding{51}& 15.2 & 38.9& 62 & 69\\
    \bottomrule
\end{tabular}
}
\label{table:Ablation_study}
\end{minipage}
\begin{minipage}{0.34\linewidth}
\captionsetup{labelfont=small}
\begin{minipage}{\linewidth}
\centering
\caption{\small Alignment method analysis.}
    \vspace{-0.2cm}
\resizebox{\linewidth}{!}{
\begin{tabular}{p{0.5\linewidth}cccc}
    \toprule
 & \multicolumn{2}{c|}{\textbf{\small ScanQA}} & \multicolumn{2}{c}{\textbf{\small R2R}} \\
 &M$\uparrow$&R$\uparrow$&SPL$\uparrow$&SR$\uparrow$\\ \hline
{\small \textit{w/o} Alignment}&14.6&35.7&56&63\\
{\small Post-Process Align}&14.9&35.2&51&59\\
{\small Train Align(our)}&15.2 & 38.9& 62 & 69\\
    \bottomrule
\end{tabular}
}
\label{table:Alignment_ana}
\end{minipage}
\begin{minipage}{\linewidth}
\caption{\small Feature-level Fusion \textit{vs.} Model-level Fusion.}
\centering
    \vspace{-0.2cm}
\resizebox{\linewidth}{!}{
\begin{tabular}{p{0.5\linewidth}cccc}
    \toprule
 & \multicolumn{2}{c|}{\textbf{\small ScanQA}} & \multicolumn{2}{c}{\textbf{\small R2R}} \\
 &M$\uparrow$&R$\uparrow$&SPL$\uparrow$&SR$\uparrow$\\ \hline
{\small Feature-level}&13.9&31.7&-&-\\
{\small Model-level(our)}&15.2 & 38.9& 62 & 69\\
    \bottomrule
\end{tabular}
}
\label{table:fusion_level}
\end{minipage}
\end{minipage}
\end{table*}
\begin{figure*}[t]
    \captionsetup{labelfont=small}
    \centering
    \includegraphics[width=1\linewidth]{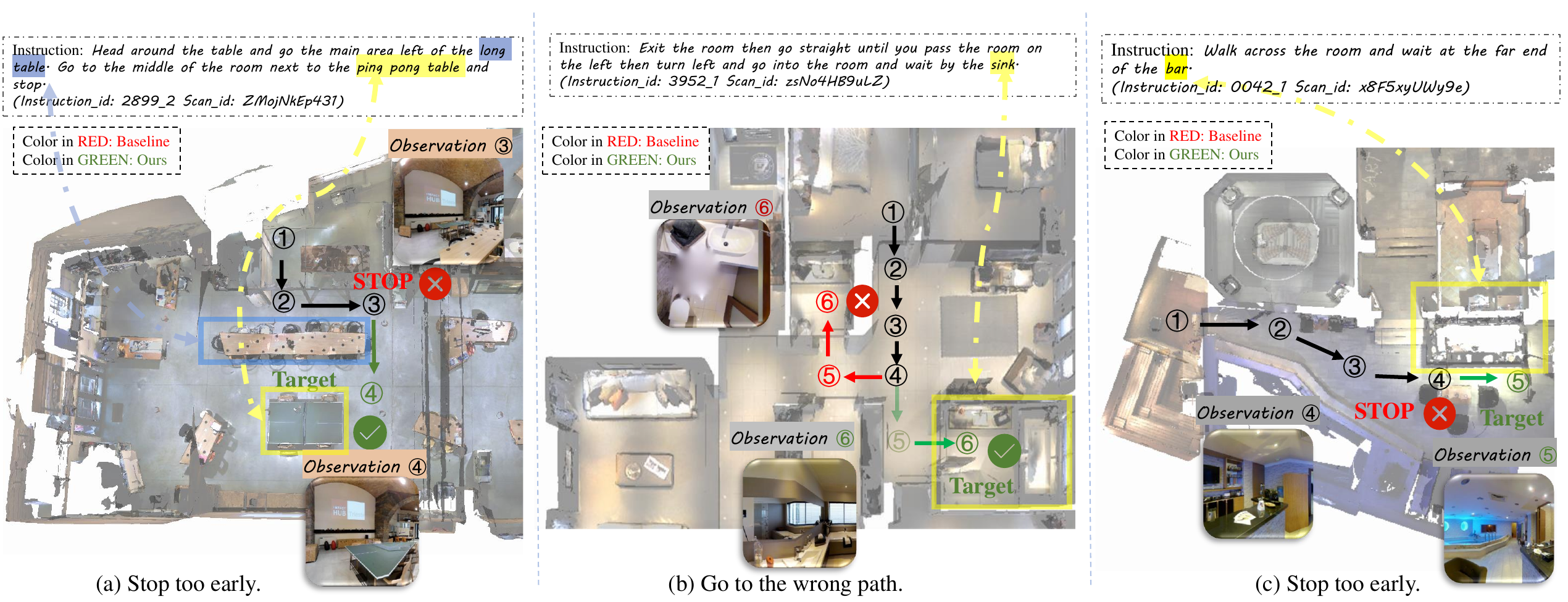} 
\caption{\small Visualization of CoNav in R2R. In Figure (a), the baseline image-text navigation agent rely on only visual cues, stops prematurely before reaching the target. This is likely caused by the image-text agent being unaware of if it is in the middle of the room, mistakenly stopping when observing the ping pong table. CoNav, on the other hand makes the correct decision. In Figure (b), there are two sinks. During navigation, the agent first encounters the incorrect sink and then see the second, correct sink. The image-text agent moves towards the first incorrect sink as soon as it observes it. This may be caused by the agent's insensitivity to directions (e.g., left/right) and its inability to determine when it has passed the left room. With Collaborative Cross-Modal Reasoning, CoNav finds the correct path. In Figure (c), the image-text navigation agent stops at the closer end of the bar without reaching the far end. This may be due to pixel-level color information not providing distance cues, whereas the 3D geometric structure can provide such distance guidance. 
}
\end{figure*}

\section{Conclusion}
\label{sec:Conclu}
In conclusion, we present a novel approach, \textbf{CoNav}, which utilizes 3D-text model guidance to enhance image-text navigation agents, addressing the challenges of fusing data from different modalities. CoNav resolves conflicting beliefs by introducing Cross-Modal Belief Alignment, effectively integrating visual cues with geometric structures during inference. This allows the agent to resolve ambiguities in embodied navigation, leveraging both visual and spatial-semantic knowledge. Our work provides a solid foundation for advancing embodied navigation tasks in real-world applications.

\clearpage

{
\small

\bibliography{main.bib}
\bibliographystyle{plain}
}

\clearpage

\appendix

\renewcommand{\KwSty}[1]{\textnormal{\textcolor{magenta!90!black}{\ttfamily\bfseries #1}}\unskip}
\newcommand{\forinline}{ \textcolor{magenta!90!black} }
\renewcommand{\ArgSty}[1]{\textnormal{\ttfamily #1}\unskip}
\SetKwComment{Comment}{\color{green!50!black}\# }{}
\renewcommand{\CommentSty}[1]{\textnormal{\ttfamily\color{green!50!black}#1}\unskip}
\newcommand{\assign}{\leftarrow}
\newcommand{\var}{\texttt}
\newcommand{\FuncCall}[2]{\texttt{\bfseries #1(#2)}}
\SetKwProg{Function}{def}{:}{}
\renewcommand{\ProgSty}[1]{\texttt{\bfseries #1}}
\SetKwProg{For}{for}{:}{}
\SetKwProg{If}{if}{:}{}
\newcommand{\VarSty}[1]{\textnormal{\ttfamily\color{blue!90!black}#1}\unskip}
\newcommand{\PredSty}[1]{\textnormal{\ttfamily\color{mygreen!90!black}#1}\unskip}
\appendix
\section*{
    \begin{center}
       Appendices and Supplementary Material of
  \resizebox{0.18\textwidth}{!}{%
    \begin{tikzpicture}[scale=1,transform shape,baseline]
      \node[anchor=base,text=cyan,opacity=1] at (0.1,-0.2) {\textbf{\Huge CoNav}};
      \node[anchor=base,text=codeblue] at (0,-0.1) {\textbf{\Huge CoNav}};
    \end{tikzpicture}%
  }%
        \end{center}
\begin{center}
\vspace{-0.3cm}
        \includegraphics[height=1.5cm]{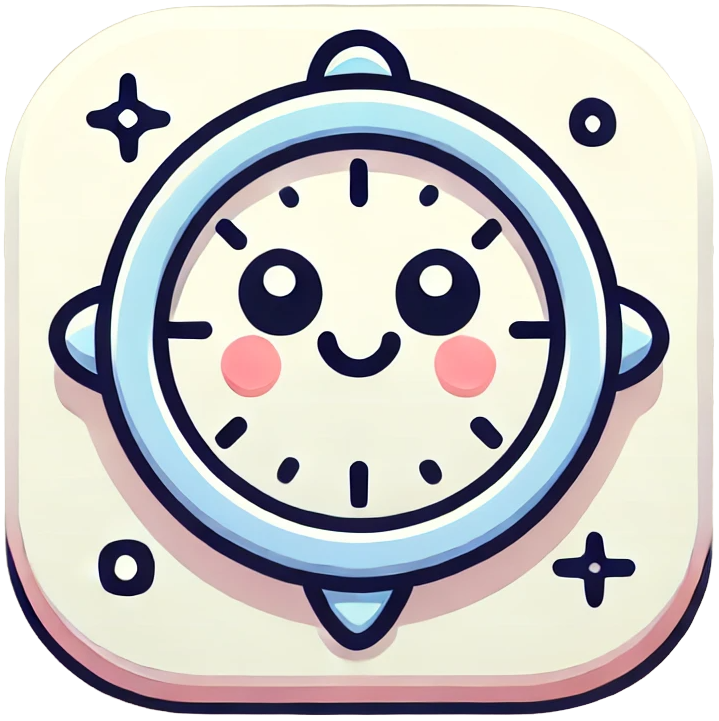}  
    \end{center}
}
\section{Implementation and Training details}
\begin{figure}[b]
    \centering
    \includegraphics[width=0.65\textwidth]{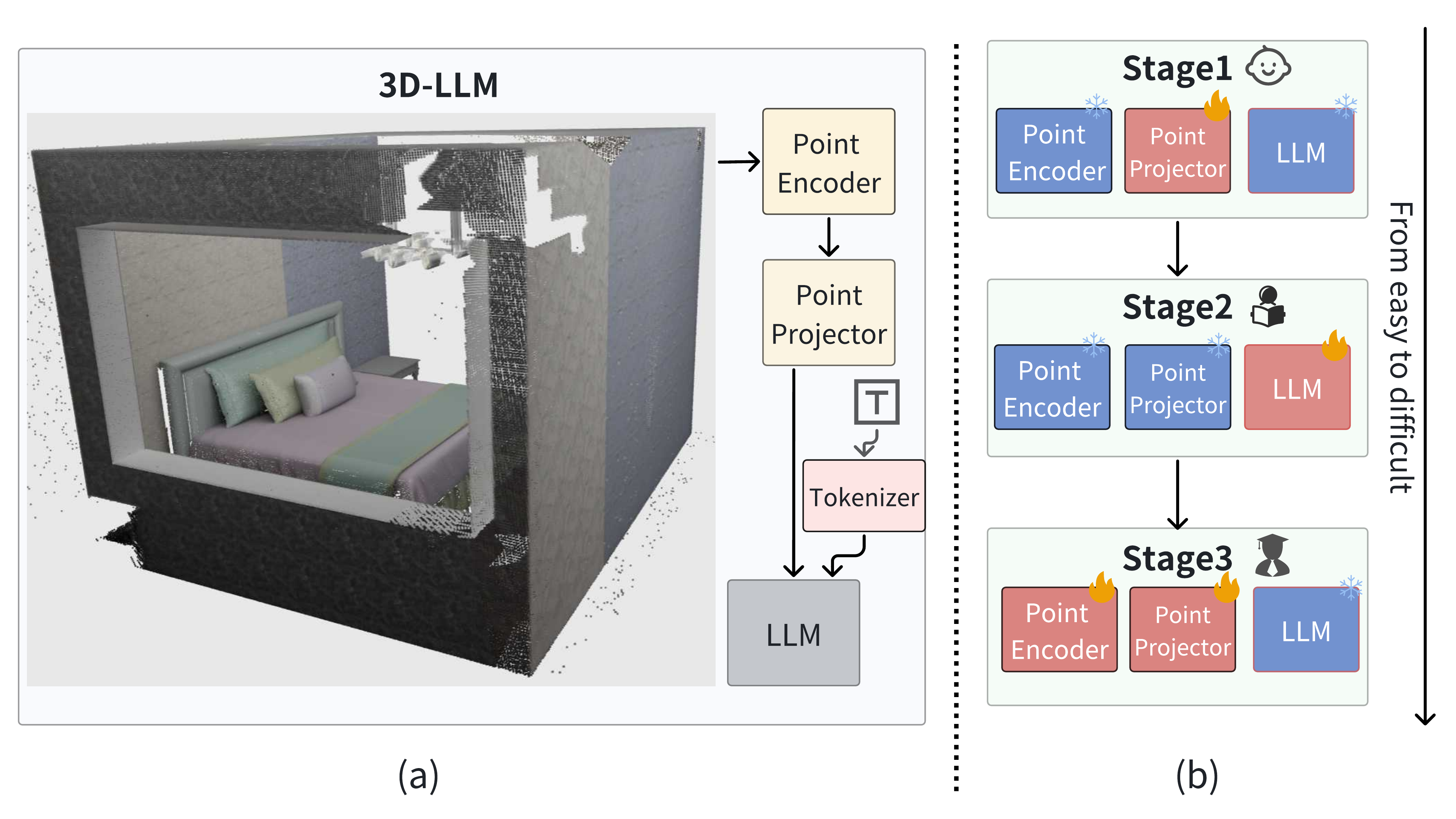}
    \caption{(a) Structure of our auxiliar 3D-text model. (b) Progressive Curriculum Learning training paradigm of our 3D-text model.}
    \label{fig:3dllm}
\end{figure}

\begin{table*}[t!]
    \caption{Qualitative results. Our 3D-text model can give reasonable answers in basic simple point clouds and complex scenes with multiple objects.}  
    \centering  
    \scalebox{0.98}{
    \scriptsize 
      \begin{tabular}{l p{0.9\linewidth} }
        \toprule
        \multicolumn{2}{l}{\bf Some Point-Text pairs used in our 3D-text model:}  \\
        \midrule
        &  \begin{minipage}[c]{0.49\linewidth}
              \centering
              \includegraphics[width=6cm,height=5cm]{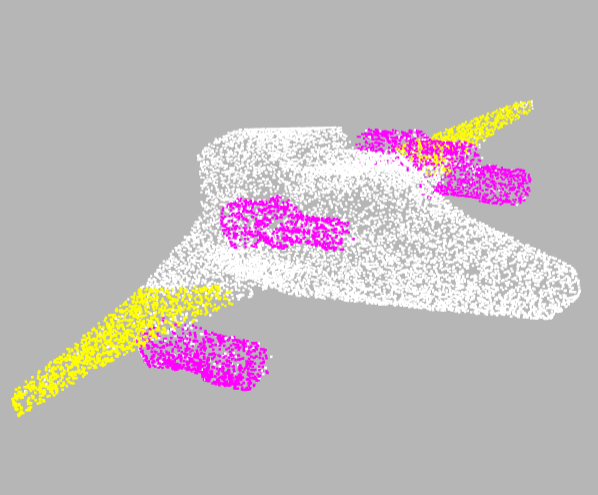} \\
              {\tiny (a) Point cloud from Cap3D in stage 1. }
           \end{minipage}
           \hfill
           \begin{minipage}[c]{0.49\linewidth}
              \centering
              \includegraphics[width=6cm,height=5cm]{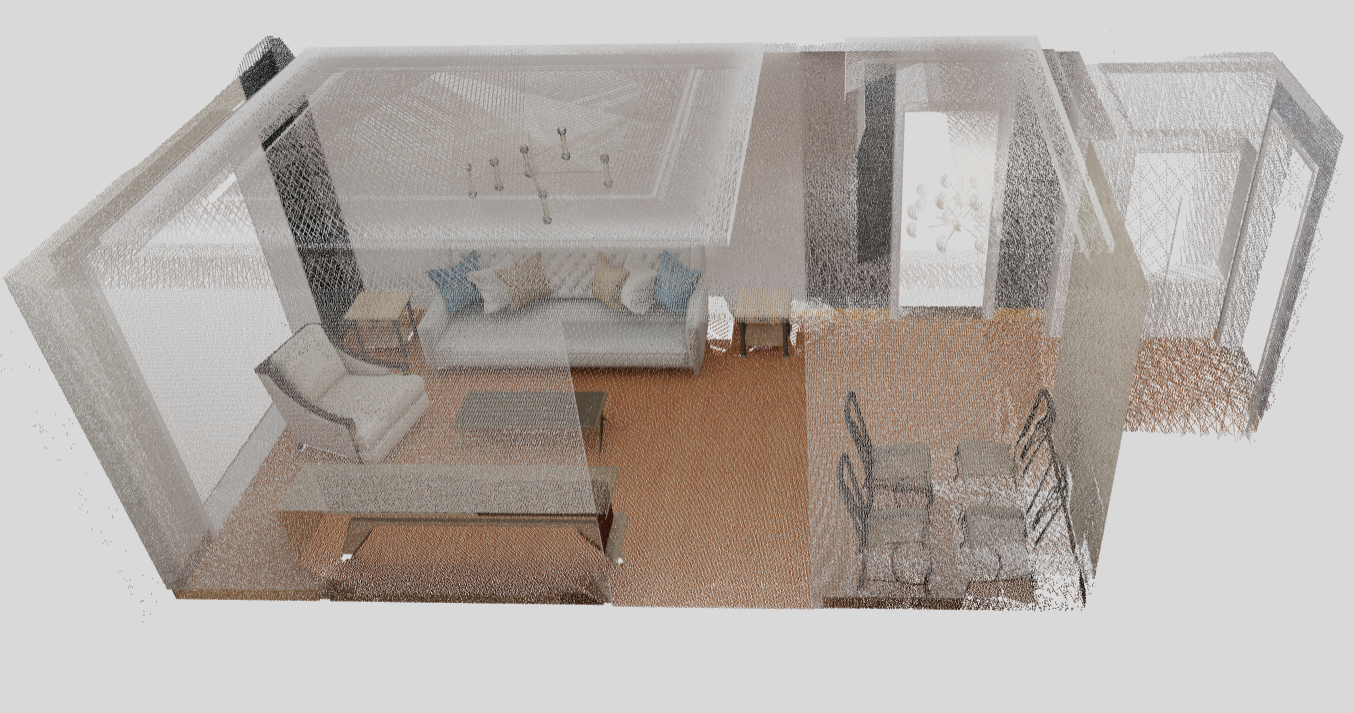} \\
              {\tiny (b) Point cloud from 3D-FRONT in stage 2. }
           \end{minipage}
           \hfill
           \begin{minipage}[c]{0.49\linewidth}
              \centering
              \includegraphics[width=6cm,height=5cm]{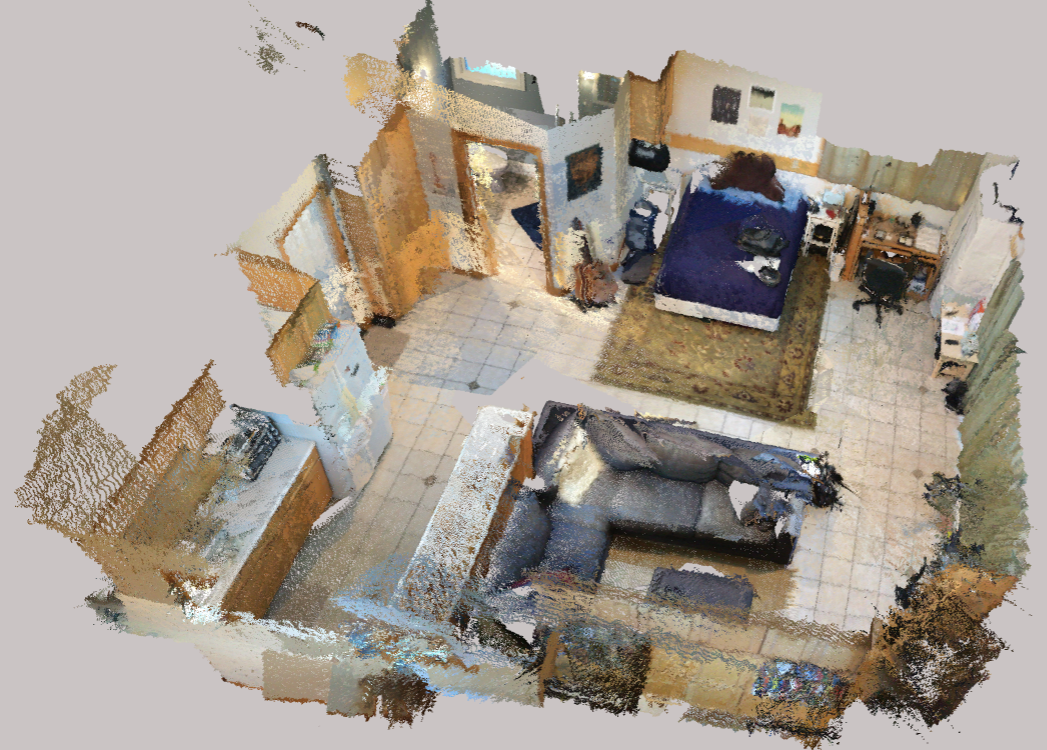} \\
              {\tiny (c) Point cloud of our dataset extracted from Scannet in stage 3. }
           \end{minipage}
           \hfill
           \begin{minipage}[c]{0.49\linewidth}
              \centering
              \includegraphics[width=6cm,height=5cm]{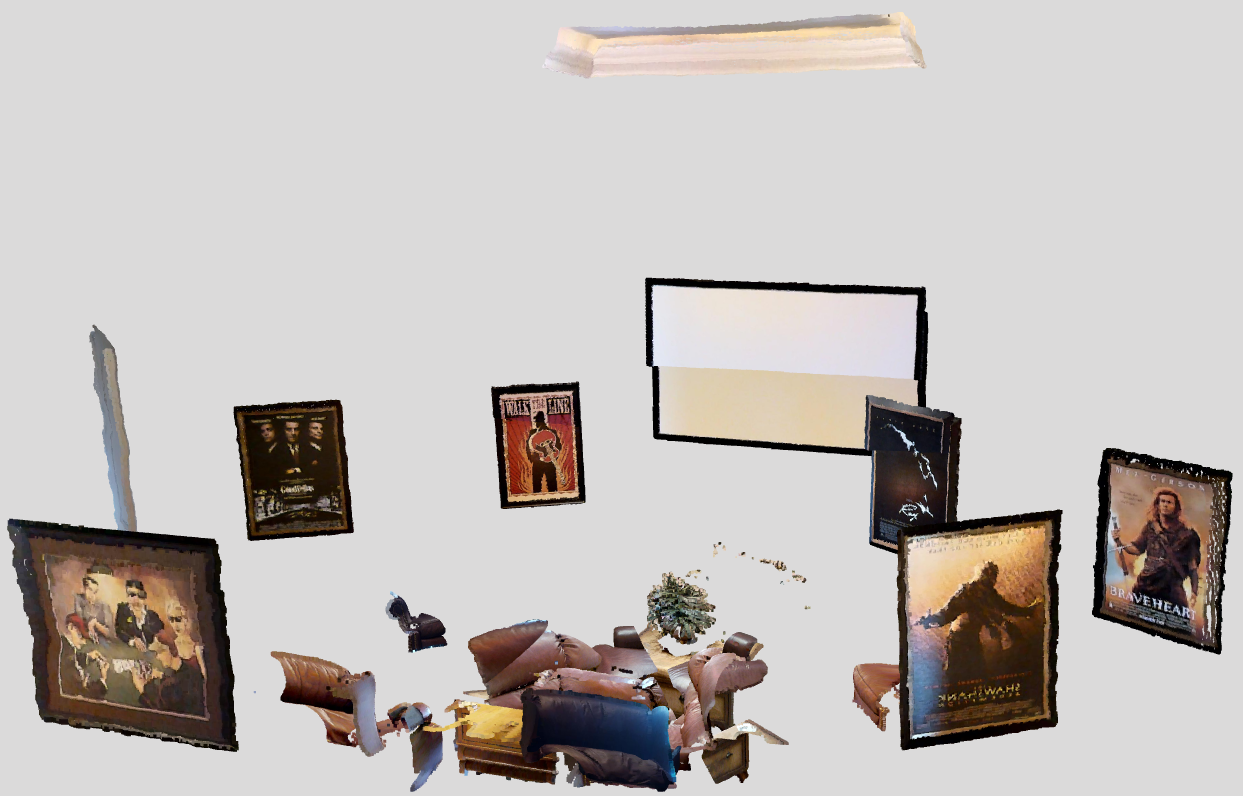} \\
              {\tiny (d) Point cloud of our dataset extracted from R2R in stage 3. }
           \end{minipage}\\
                   \midrule
\small 
\scriptsize
        Text & (a) Summarize the 3D point cloud object briefly.  \\
        PointLLM   & The 3D model represents a bright, cartoon-style airplane, predominantly white with contrasting yellow wings. The design, simplified and stylized, appears child-friendly, suggesting its usage as a toy or a decorative item in a children's room. \\
Ours stage 2 & A playful toy plane, predominantly in white color, is adorned with vibrant yellow wings that make it stand out. The design of this object implies it is meant for children to enjoy and learn from, perhaps by engaging in imaginative play or study of aeronautics. \\
Ours stage 3 & The 3D object model airplane has a colorful design, with the wings featuring a mix of bright yellow and pink hues and the body being white \\
\midrule
        Text & (b) How would you interpret this 3D point cloud in detail?  \\
        PointLLM   & This 3D point cloud presents an interior layout of a house with a cartoon-like aesthetic, emphasizing simplicity and minimalism. \\
                Ours stage 2 & In this room, the blend of diverse design elements creates a harmonious and functional living environment. The presence of the dining table and dining chair suggests a space dedicated to meals and community gatherings. The desk is conveniently located for study or work... \\
                Ours stage 3 &This 3D point cloud represents a living area in a house with a Neoclassical style, characterized by its luxurious aesthetic. The corner table serves as a versatile piece with its warm brown color and wooden texture, enhancing the comfort of the room. The pendant lamp hangs above the central area of the room...\\
        \midrule
        Text & (c) What is the spatial relationship between the bed and the sofa in this point cloud?  \\
              PointLLM   &The relationship between the bed and the sofa in front of a large window is one of visual connection, offering an inner view of the room. \\   
        Ours stage 2 &The sofa is placed closely adjacent to the bed, suggesting a common living area in a shared room. \\
        Ours stage 3 & The sofa and the bed are adjacent to each other, separated by a minimal distance, indicating a bedroom setting designed for rest.\\ 
                \midrule
        Text & (d) How many posters are there around the sofa in this home theater? \\
              Ground truth & There are 6 posters around the sofa in the home theater.\\  
        PointLLM   & This is a broken point cloud without posters. \\ 
        Ours stage 2 & There are six poster art pieces around the sofa in this home theater.\\
        Ours stage 3 & There are six posters in the home theater.\\
        \bottomrule
      \end{tabular}
    }

    \label{tab:qualitative_results}  
\end{table*} 

\subsection{Training details of 3D-text model}
\label{sec:append_3dllm}
The structure of our 3D-text model is shown in Figure \ref{fig:3dllm} (a), it contains a point cloud encoder, a point cloud projector, and a LLM. The pretrained Transformer-based \textbf{point cloud encoder} encodes the point cloud $p$. First the encoder samples $m$ center points from the point cloud, then applies a k-nearest neighbor algorithm to select the nearest neighbors for each center point, grouping $m$ local patches. The encoder outputs a sequence of point features, denoted as $G = (g_1, g_2, \dots, g_m) \in \mathbb{R}^{m \times c}$, where $m$ is the number of point features and $c$ is the feature dimension. The \textbf{point cloud projector} maps the point features $G$ to the word token space. The \textbf{pre-trained LLM} backbone processes point cloud $p$ and text tokens $t$. During pretraining, point $p$ and text $t$ come from Point-Text pairs indataset, while during Cross-Modal Belief Alignment fine-tuning and inference, they come from the small triple‐modality corpus \(\mathcal{D}_{\mathrm{3MT}}\).

Inspired by \cite{xu2025pointllm, zhou2025navgpt}, we carefully design tasks with corresponding datasets from simple to complex and real-world and divide the training into three different stages. At the same time, we carefully arrange different frozen and trainable components distributions in 3D-text model to balance performance and efficiency as follows:
\begin{equation}
\begin{aligned}
  \pi_{3D}(p, L) 
  &= \mathrm{LLM}^{\text{\TFicon{F}/\TFicon{T}}}
     \Bigl(
       \mathrm{Proj}^{\text{\TFicon{F}/\TFicon{T}}}
       \bigl(\mathrm{Enc}^{\text{\TFicon{F}/\TFicon{T}}}(p)\bigr),
       L
     \Bigr),\\
  \min_{\theta_{3D}}\ 
  &\mathbb{E}_{(p, L, \hat a_t)\sim\mathcal{D}_{3D}}
    \Bigl[\,
      \mathcal{L}\bigl(\pi_{3D}(p,L),\,\hat a_t\bigr)
    \Bigr].
\end{aligned}
\end{equation}
Here, \TFicon{F} / \TFicon{T} represents whether the component (e.g., the point encoder, projector, or LLM) is frozen (\TFicon{F}) or trainable (\TFicon{T}) at this stage.  \(\pi_{\text{3D}}\) represents the image-text navigation agent’s action. The term \(\mathcal{D}_\text{3D}\) refers to the dataset in the current stage, and \(\mathcal{L}\) is the loss function used to optimize the model. \(\hat{a_t}\) is the ground truth action. The progression from Stage 1 (simple tasks) to Stage 3 (real-world data) enables the model to incrementally learn more complex spatial relationships and descriptions, while adapting to the increasing difficulty of the task. 

The summary of the Curriculum Learning schedule is shown in Table \ref{table:stages}.
Inspired by \cite{xu2025pointllm, zhou2025navgpt}, we believe that the key to enabling 3D-text model to handle complex real-world tasks lies in the design of the dataset and the balance between frozen and trainable components. Therefore, we divide the tasks and the corresponding datasets into three stages based on difficulty, from simple to complex and real-world, as shown in Figure \ref{fig:3dllm} (b). \textbf{Stage 1}. For simple tasks, our goal is to align point clouds with LLM. Since the point encoder and LLM are pretrained, inspired by LLava, we \textit{freeze} the point encoder and LLM and \textit{train} the point projector. We use the 770k data from Cap3D \cite{Cap3D}, annotated by \cite{xu2025pointllm}. This dataset contains a large number of point clouds of individual objects, covering various types of 3D object, and is used for simple description tasks. 
\textbf{Stage 2}. In complex tasks, our goal is to enable 3D-text model to handle indoor scenes with multiple objects and describe the spatial relationships between them. We use the 3D-FRONT \cite{fu20213d} dataset, annotated by 3D-GRAND \cite{yang20243d}. The 3D-FRONT dataset contains high-quality synthetic 3D indoor scenes, such as living rooms, bedrooms, and kitchens. We complete tasks like Detailed Description of the scene and Spatial Relation QA on it. Since we already aligned point clouds with LLM in Stage 1, we freeze the point encoder and the point projector here. This stage focuses on learning multi-object scene descriptions and spatial reasoning, which were not addressed in Stage 1. Therefore, we \textit{freeze} the point encoder, the point projector, and  \textit{train}  the LLM.
After Stage 2, the 3D-text model can handle multi-object scenes and more complex spatial reasoning. However, real scanned point cloud datasets, which contain partial data from certain angles, still differ significantly from synthetic datasets. Directly applying the Stage 2 model could result in outputs like \textit{"This is a damaged point cloud"} or \textit{"The point cloud cannot be recognized."} This issue arises because the point encoder cannot properly process partial point data. To preserve the capabilities gained in Stages 1 and 2 while enabling the model to handle real scanned point clouds, \textbf{Stage 3} we \textit{freeze }the LLM and \textit{train} the point encoder and projector using Point-Text annotations generated by our pipeline from R2R \cite{anderson2018r2r} and ScanNet \cite{azuma2022scanqa}. We apply 3D QA and Brief Bescription tasks in Stage 3 to train the 3D-text model.

\begin{table*}[tb]
\centering
\caption{Overview of the Curriculum Learning schedule for 3D-text model. \textbf{\textcolor{blue}{\ding{55}}} stands for frozen and \textbf{\textcolor{red}{\ding{51}}} stands for trainable.}
\resizebox{0.9\textwidth}{!}{
\begin{tabular}{c|c|c|c|ccc|c} 
    \toprule
\multirow{2}{*}{\textbf{Stage}} & \multirow{2}{*}{\textbf{Goal}} & \multirow{2}{*}{\textbf{Dataset}} & \multirow{2}{*}{\textbf{Task}} & \multicolumn{3}{c|}{\textbf{Trainable Components}} & \multirow{2}{*}{\textbf{Example Instruction}} \\ 
                                &                                     &                                   &                                 & \small Encoder & Projector & LLM & \\ \hline
 1 simple & Align point clouds with LLM.       & Cap3D        & Brief Description        &     \textbf{\textcolor{blue}{\ding{55}}}       & \textbf{\textcolor{red}{\ding{51}}}        & \textbf{\textcolor{blue}{\ding{55}}}  & \parbox[c][0.8cm][c]{4cm}{\small"How would you interpret this 3D point cloud?"} \\ \hline
\multirow{2}{*}{ 2 complex} & Enable 3D-text model to handle complex & \multirow{2}{*}{3D-FRONT}   & Detailed Description        &     \multirow{2}{*}{\textbf{\textcolor{blue}{\ding{55}}}  }     &   \multirow{2}{*}{\textbf{\textcolor{blue}{\ding{55}}}}     & \multirow{2}{*}{\textbf{\textcolor{red}{\ding{51}}}}   & \parbox[c][0.8cm][c]{4cm}{\small"Could you elaborate extensively on what this represents?"} \\ 
                                 & multiple objects indoor scenes. & &Spatial Relation QA&  & & & \parbox[c][0.8cm][c]{4cm}{\small"What is the spatial relationship between TV and sofa?"}\\ \hline
\multirow{2}{*}{ 3 real-world} & Maintain spatial reasoning capabilities. & R2R point    & Brief Description        &   \multirow{2}{*}{\textbf{\textcolor{red}{\ding{51}}}   }    & \multirow{2}{*}{\textbf{\textcolor{red}{\ding{51}}}  }    &   \multirow{2}{*}{\textbf{\textcolor{blue}{\ding{55}}}  }& \parbox[c][0.8cm][c]{4cm}{\small"What detailed insights can you give about this point cloud?"} \\ 
                                 & Adapt to scanned point cloud data. &ScanQA &3D QA&  & & & \parbox[c][0.8cm][c]{4cm}{\small"What is placed next to the fridge upper of the cabinets?"}\\     \bottomrule
\end{tabular}
}
\label{table:stages}
\end{table*}

\begin{table}[tb]
    \centering   
    \caption{Overview of data in different stages. GT stands for ground truth. Instr stands for instruction. "mesh" means the point cloud generated from 3D mesh and "scan" means the point cloud generated by camera scan. Pairs stands for the number of Point-Text pairs in this stage. }
    \scriptsize
    \resizebox{0.75\textwidth}{!}{
    \begin{tabular}{p{0.01\textwidth}|p{0.16\textwidth}|p{0.16\textwidth}|p{0.16\textwidth}}
    \toprule
    \textbf{\rotatebox{90}{Scene}} & 
          \begin{minipage}{0.16\textwidth}
           \includegraphics[width=\textwidth,height=0.75\textwidth]{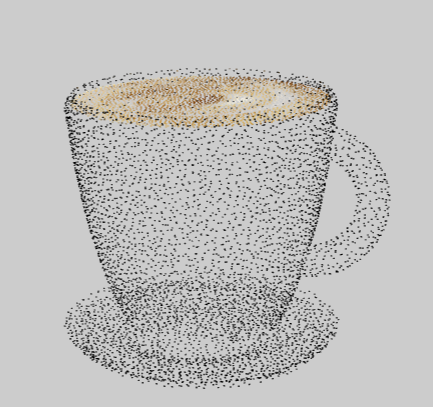} \\
    {\tiny (a) Point cloud in Stage 1.}
\end{minipage} &
           \begin{minipage}{0.16\textwidth}
           \includegraphics[width=\textwidth,height=0.75\textwidth]{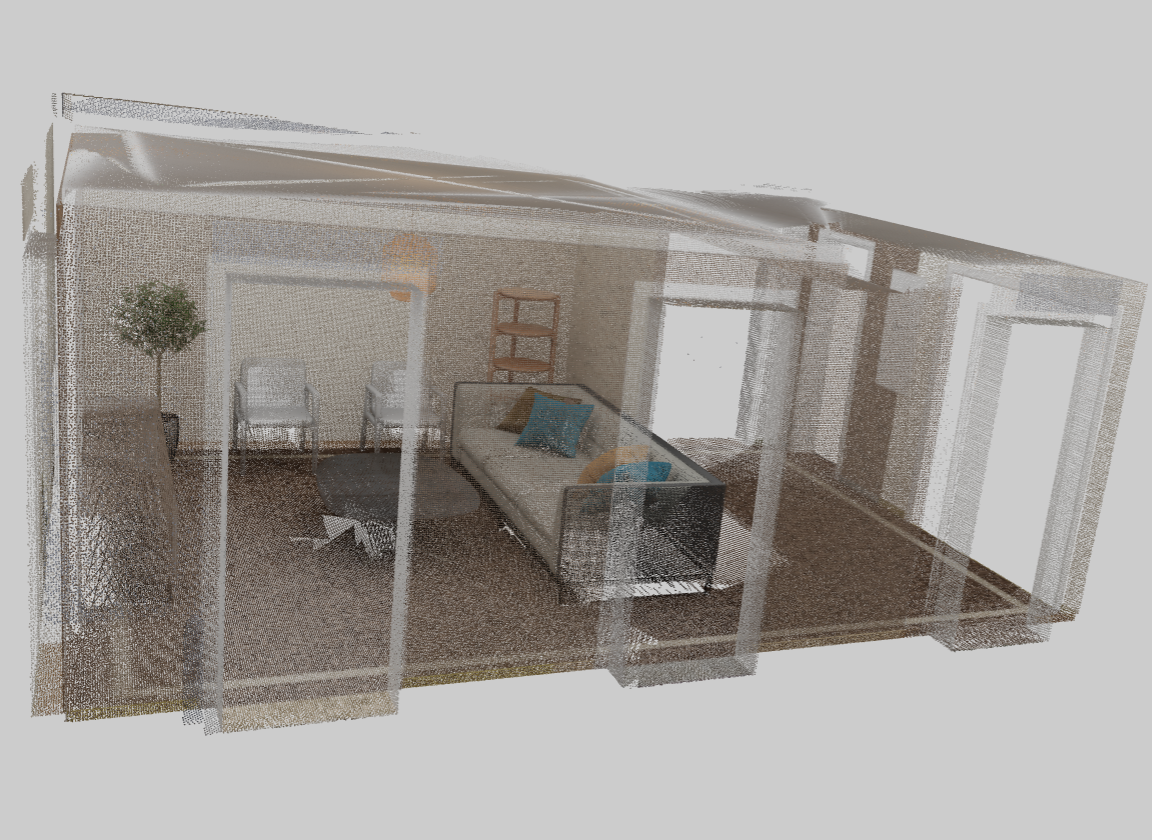} \\
    {\tiny (b) Point cloud in Stage 2.}
\end{minipage} &
          \begin{minipage}{0.16\textwidth}
           \includegraphics[width=\textwidth,height=0.75\textwidth]{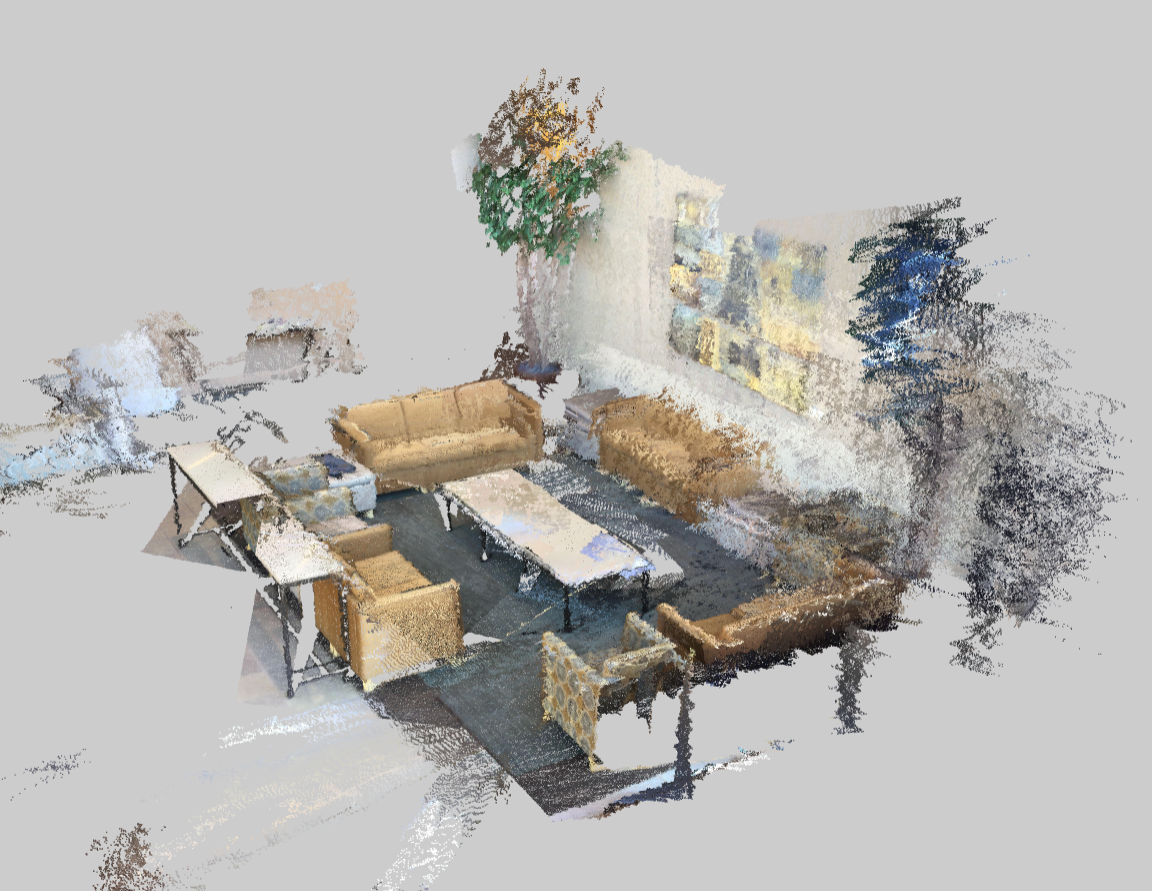} \\
    {\tiny (c) Point cloud in Stage 3.}
\end{minipage}  
    \\ \midrule
    \textbf{\rotatebox{90}{Instr}} & \begin{minipage}{0.15\textwidth}\centering Give a brief explanation of the object that this cloud of points forms. \end{minipage}&\begin{minipage}{0.15\textwidth}\centering  What's the spatial relationship between the tv stand and three-seat sofa?\end{minipage}& \begin{minipage}{0.15\textwidth}\centering What color leather sofa is in the room?\end{minipage}\\ \midrule
    \textbf{\rotatebox{90}{GT}} & \begin{minipage}{0.15\textwidth}\centering Black coffee cup and saucer 3D model. \end{minipage}&\begin{minipage}{0.15\textwidth}\centering Tv stand is far from the three-seat sofa.\end{minipage}& \begin{minipage}{0.14\textwidth}\centering  brown\end{minipage}\\ \midrule
    \textbf{\rotatebox{90}{Type}} & \begin{minipage}{0.15\textwidth}\centering \textbf{mesh }\end{minipage}&\begin{minipage}{0.15\textwidth}\centering \textbf{mesh}\end{minipage}& \begin{minipage}{0.15\textwidth}\centering \textbf{scan}\end{minipage}\\ \midrule
    \textbf{\rotatebox{90}{Pair}} & \begin{minipage}{0.15\textwidth}\centering 770K \end{minipage}&\begin{minipage}{0.15\textwidth}\centering 51K\end{minipage}& \begin{minipage}{0.14\textwidth}\centering 73K\end{minipage}\\
    \bottomrule
    \end{tabular}
    }
    \label{tab:examp}
\end{table}

\subsection{Training times and hyperparameters for training}
\label{sec:append_hyperparameters}
Training times for each stage and the navigation model are summarized in Table \ref{table:training_time}. All reported values are converted to 1*A100 GPU‐hours.
We train the image–text navigation model on diverse embodied navigation image–text pairs, fully following the approach in NaviLLM \cite{zheng2023towards}. Therefore, we omit the details of the hyperparameters used for training the image–text navigation model.
The hyperparameters for training 3D-text model are presented in Table \ref{tab:append_param_all}. 
The hyperparameters for Cross-Modal Belief Alignment in Table \ref{tab:append_param_nav}.
The hyperparameters for inference are listed in Table \ref{tab:append_param_inf}.
\subsection{Qualitative Result of our 3D-text model.} 
We present a visual comparison of our 3D-LLM at different stages with other 3D-LLM methods in Table \ref{tab:qualitative_results}. Our 3D-LLM demonstrates superior capability in managing complex real-world scenes that contain multiple objects, while still maintaining proficiency in the description and analysis of basic simple object point cloud. 

\subsection{Effectiveness of our Curriculum Learning schedule.} 
The data presented in rows 1 to 4 in Table \ref{tab:append_ab_exp_}, the improvement in performance illustrates the absence of the Curriculum Learning schedule impairs the spatial reasoning ability in complex multi-object real-world scenes, which is important in Embodied Navigation. 
\begin{table}[bth]
\caption{Ablation study of Curriculum Learning schedule. }
\centering

\begin{tabular}{ccc|ccc|cc|cc}
    \toprule
 \multicolumn{3}{c|} {\textbf{Stage}} & \multicolumn{3}{c|} {\textbf{Modality}} & \multicolumn{2}{c|}{\textbf{ScanQA}} & \multicolumn{2}{c}{\textbf{R2R}} \\
1&2&3&Text&Image&Point cloud&M&R&SPL&SR\\ \hline
\ding{55} & \ding{55} & \ding{55} & \ding{51} & \ding{51}& \ding{55}& 14.3  & 36.7&57&64 \\
\ding{51} & \ding{55} & \ding{55} & \ding{51} & \ding{55}& \ding{51} & 8.39 & 7.89&- & -\\
\ding{51} & \ding{51} & \ding{55} & \ding{51} & \ding{55}& \ding{51} & 10.1 & 9.78  &  - & -\\
\ding{51} & \ding{51} & \ding{51}& \ding{51} & \ding{55}& \ding{51} & 14.5 & 37.1 & - & -\\
\ding{51} & \ding{51} & \ding{51} & \ding{51} & \ding{51}& \ding{51} & 15.2 & 38.9& 62 & 69\\
    \bottomrule
\end{tabular}

\label{tab:append_ab_exp_}  
\end{table}
\begin{figure}[tbp]
    \centering
    \includegraphics[width=0.85\textwidth]{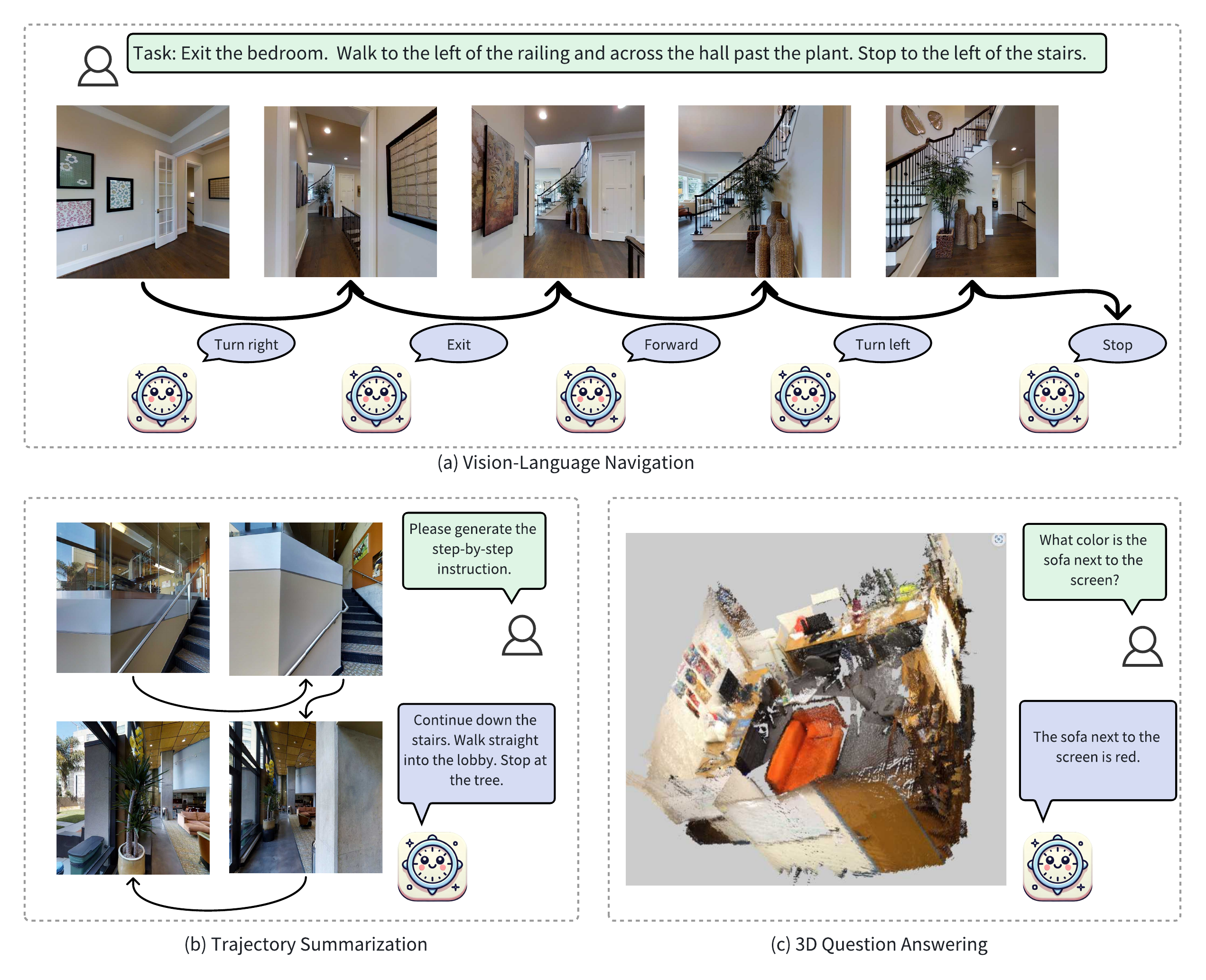} 
    \caption{More Qualitative results of CoNav for different Embodied Navigation task.}
    \label{fig:appendix_data_pipline}
\end{figure}

\section{Data Collection in Stage 3}
\subsection{Our Point-Text pairs generation pipline in Stage 3 for 3D-text model.}
    \label{sec:appendix_data_pipline}
    
    \begin{figure}[H]
    \centering
    \includegraphics[width=\linewidth]{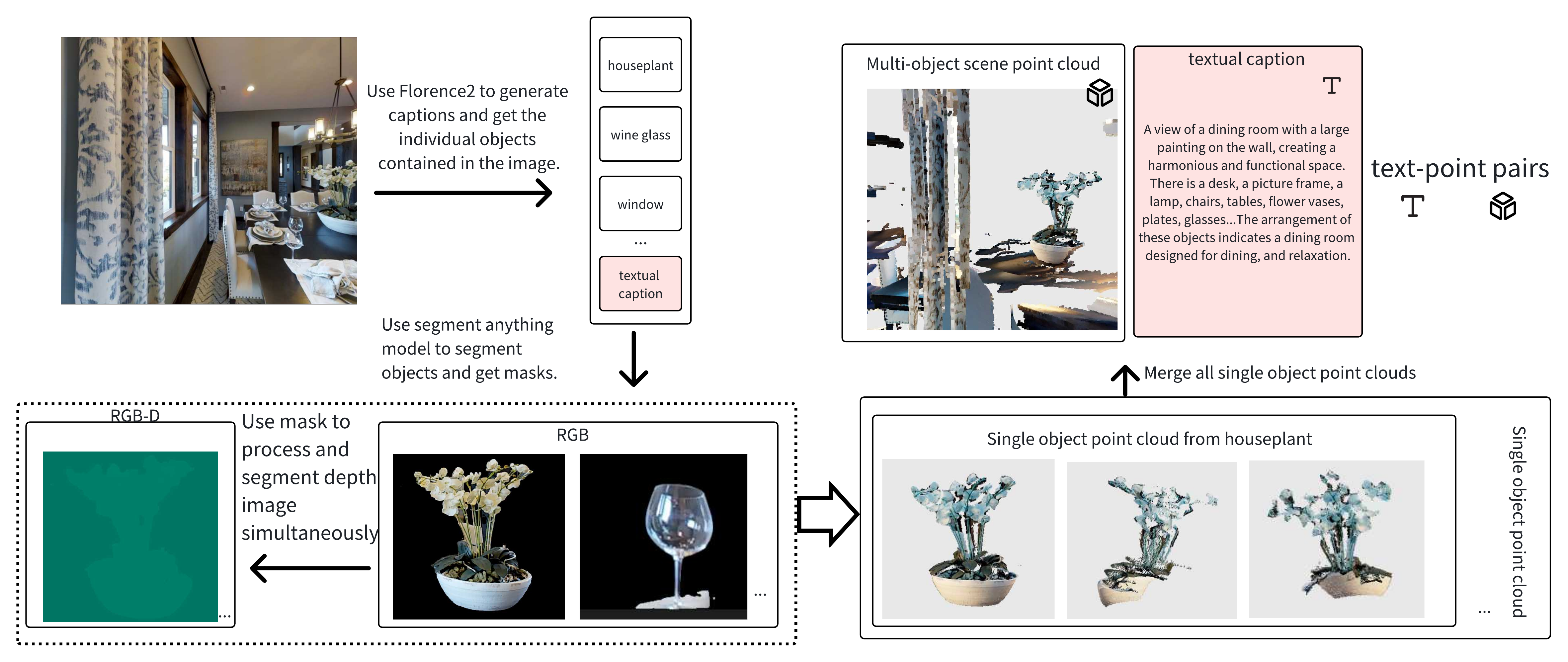} 
    \caption{Our Point-Text pairs generation pipline in Stage 3 for pretraining 3D-text model.}
    \label{fig:appendix_data_pipline}
\end{figure}

 In the Curriculum Learning schedule, in Stage 1 we directly use Cap3D \cite{Cap3D}, annotated by \cite{xu2025pointllm}, and in Stage 2 we directly use the 3D-FRONT \cite{fu20213d} dataset annotated by 3D-GRAND \cite{yang20243d}. However, R2R and ScanQA do not have suitable Point-Text pairs, so we refer to \cite{chen2024spatialvlm} and design a data generation pipeline, as shown in Figure \ref{fig:appendix_data_pipline}.\\
The Point-Text annotations used in Stage 3 is extracted directly by RGB images and depth images, combined with camera internal and external parameters (detailed in the Appendix \ref{sec:appendix_data_pipline}). We refer to \cite{chen2024spatialvlm}, first use Florence2 \cite{xiao2024florence} to identify the existing objects, then use SAM \cite{kirillov2023segment} to segment the identified objects from RGB and RGB-D one by one to form a point cloud of a single object, finally combine all the single object point clouds to form a point cloud of the complete scene. Here are some examples of different stages in Table \ref{tab:examp}, you can find more in Table \ref{tab:append_more_visualization_data}. 
 In R2R, we refer to \cite{chen2024spatialvlm}, first use Florence2 \cite{xiao2024florence} to identify the existing objects. At the same time, the caption of the entire scene is obtained, as shown in the figure. Note that for clarity, the scene RGB image in the figure only captures a small part. In fact, the scene graph is a 360-degree panoramic image and has a corresponding depth map (RGB-D). Then use SAM \cite{kirillov2023segment} to segment the identified objects from RGB and RGB-D one by one to form a point cloud of a single object. Because the scanned image of the R2R camera is radial from the center to the surroundings, if it is not segmented and the debris is not removed, the point cloud will be chaotic and difficult to identify. If ScanQA does not scan from the center to the surroundings, it does not need to segment individual objects. Finally combine all the single object point clouds to form a point cloud of the complete scene. We collected 37 text-3D point cloud pairs (R2R contains 16k pairs of 8k scenes, Scannet contains 57k pairs of 1.5k scenes) for stage 3 in total.
\subsection{Dataset distribution}
\begin{figure*}[p]
    \centering
    \includegraphics[width=1\linewidth]{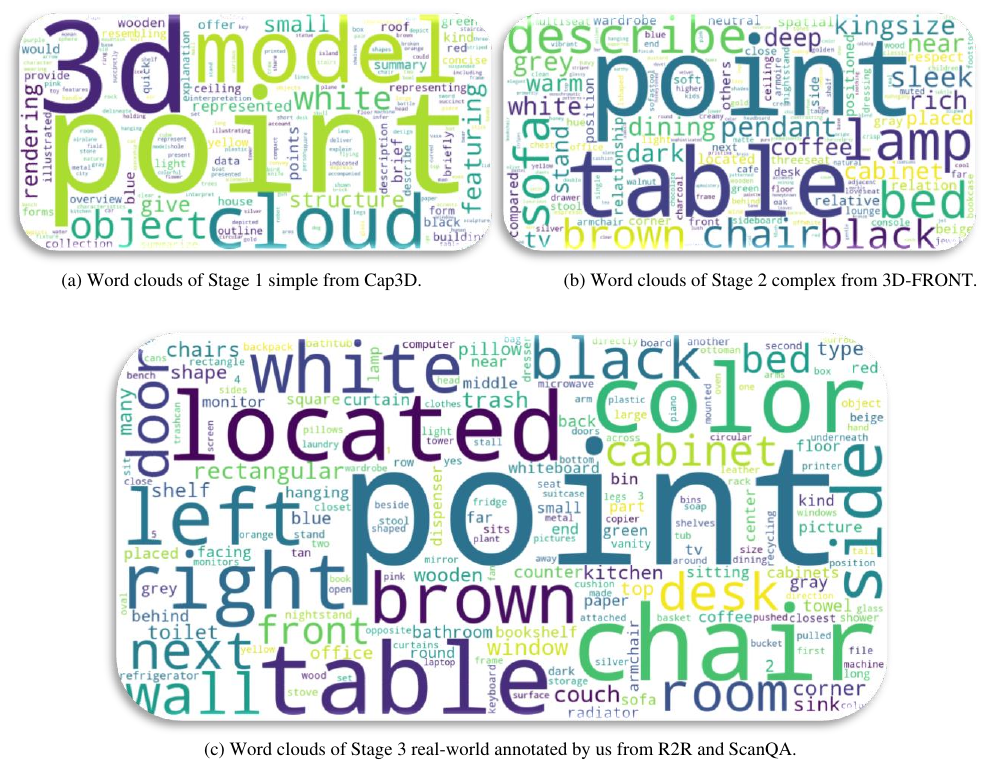} 
    \caption{Word clouds of different Stage during training of our 3D-text model.}
    \label{fig:append_Wordclouds}
\end{figure*}

\begin{figure*}[t]
    \centering
    \includegraphics[width=1\linewidth]{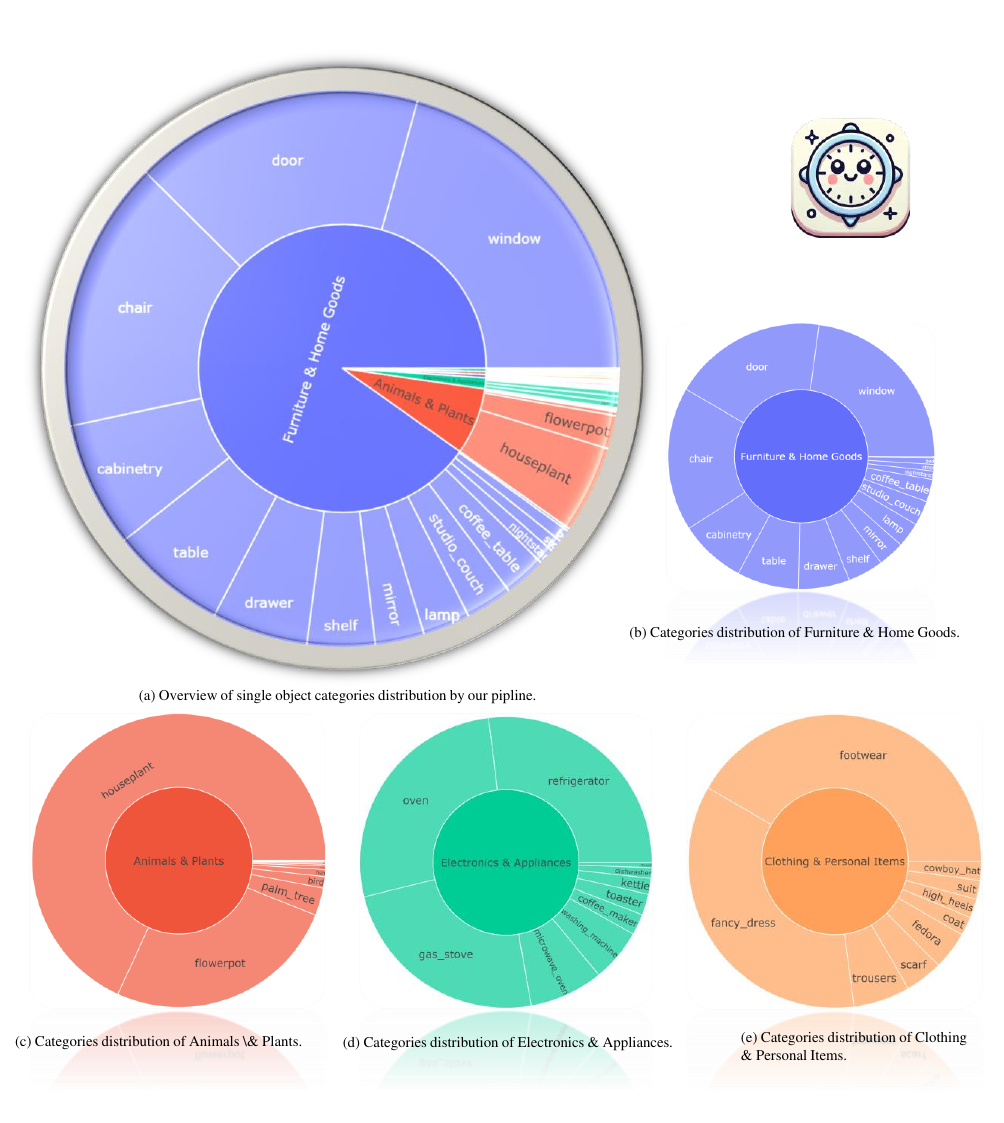} 
    \caption{Single object point cloud data category distribution.}
    \label{fig:append_categorydistribution}
\end{figure*}

For effective training, we collect a large number of text-3D point cloud pairs, covering simple Stage 1 (770k), complex Stage 2 (51k), and real-world Stage 3 tasks (73k). Figure \ref{fig:append_Wordclouds} we present word clouds of different Stages. Figure \ref{fig:append_categorydistribution} illustrates single object point cloud data category distribution generated by our pipline from R2R. Here we divide the point cloud of a single object into 6 categories, respectively Furniture \& Home Goods, Electronics \& Appliances, Entertainment \& Leisure Items, Outdoor \& Sporting Goods, Animals \& Plants, Clothing \& Personal Items, for easy display.
\subsection{Data and Task example}
\label{sec:append_Task_example}
\textbf{Data example:} In order to have a more intuitive understanding of the point cloud data, more data from different stages are shown in Table \ref{tab:append_more_visualization_data}. It can be seen intuitively from the table that in the simple split of the point cloud, the objects are single and the spatial relationship is simple. However, in the complex stage, there are more objects and the relationship is more complicated. In the real-world scene, there will be many types of objects and the point cloud is not complete. This is because in the real scene, the camera scan cannot cover every angle.
\\
\textbf{Task example:} In Table \ref{tab:append_task_example1}, \ref{tab:append_task_example2}, \ref{tab:append_task_example3}, \ref{tab:append_task_example4}, \ref{tab:append_task_example5}, we show examples of different tasks at different stages of training 3D-text model. Through this Curriculum Learning schedule, point clouds gradually transition from simple ones generated by a complete model to complex ones in real scanned scenes. The tasks gradually change from simple tasks to difficult and complex tasks. The number of objects in the scene also gradually changes from a single object to multiple objects. Just like humans learn from simple to difficult, in this process the model gradually learns 3D reasoning and understanding capabilities in real multi-object and incomplete scenes.
We use GPT-4 to generate an instruction list and then randomly select the text and point cloud generated above to form the final Point-Text cloud pairs.

\setlength{\textfloatsep}{2pt} 
\setlength{\floatsep}{2pt}     
\setlength{\intextsep}{2pt}    
\section{Broader Impact}
\label{sec:Impact}

\subsection{Limitations and Ethical Considerations.}
\textbf{Limitations}
One key limitation of our approach is that the 3D information described through text still suffers from loss and noise when compared to the original 3D geometric structure. While text-based descriptions provide valuable contextual information, they cannot fully capture the intricacies of the 3D world. Additionally, real-world scanned point cloud data can be highly resource-intensive, requiring significant computational power and memory. To mitigate these resource demands, we performed downsampling when inputting the point cloud into the point cloud encoder. However, this downsampling can introduce noise into the data, which may affect the quality of spatial reasoning. Furthermore, the data generation process sometimes depends on image-text models, which may introduce biases from the image-text model into the final trained 3D-text model, potentially influencing the model’s reasoning and decision-making.

\textbf{Ethical Considerations.}
As with any system involving data collection and processing, we must ensure that any images or point clouds used in our method do not involve personal privacy or violate ethical standards. All data used in our system must adhere to strict privacy policies to protect individuals' sensitive information. Additionally, we are committed to long-term maintenance and transparency, ensuring that user data is protected and handled responsibly. Our goal is to continually improve privacy safeguards, ensuring that our approach remains aligned with ethical principles and regulatory standards.

\textbf{Scope of Conclusions.}
It is important to acknowledge that the experiments and data, including those presented in our work, may only represent a subset of the broader range of real-world scenarios. Although our findings are based on specific datasets, including 3D structures and 2D-3D-text triples, we believe that the conclusions we have drawn provide a robust foundation for understanding indoor embodied navigation. Our results are relevant to indoor environments and provide valuable insights into how multi-modal reasoning can be leveraged for navigation tasks. However, it is essential to recognize the limitations of our datasets and the need for further research and validation in more diverse settings. Nevertheless, we believe that the principles and methodologies developed in this work are applicable to the broader field of embodied navigation, and they lay the groundwork for future advancements in this area.

\subsection{Usage of Language Models and Simulators}
We have used the LLaMA \cite{touvron2023llama} model from Meta, the Qwen2.5-7B model \cite{yang2024qwen2}, and MatterPort3D \cite{chang2017matterport3d} data, all of which are authorized for research purposes. We express our deep respect for the contributions of the developers and researchers who have made these models and datasets available. Their work has significantly advanced the field of language modeling and 3D data collection, providing the foundation for our research and enabling us to build upon their efforts to further push the boundaries of embodied navigation and multi-modal reasoning.

\section{Prompt}
\label{sec:appendix_prompt}
\subsection{Prompt format in  Communication Interface}

\begin{table}[htbp]
\centering
 \caption{Prompt format overview in our \textit{Navigation Communication Interface}. \textcolor{codeblue}{Text in codeblue means the variable in the prompt.} It organizes prompts and handles communication, providing interface for various downstream models and tasks.}
\begin{tcolorbox}[colback=lightgray!30, colframe=black, sharp corners=all, width=0.90\textwidth,  title=Communication Interface.]
\small 
\scriptsize
    \textbf{Prompt send to 3D-text model}\\
\verb|<|\textcolor{codeblue}{Chat Template}\verb|>| \verb|<|\textcolor{codeblue}{Point}\verb|>| Here are some additional task descriptions: \verb|<|\textcolor{codeblue}{Task Info}\verb|>| Here is your task: Illustrate the spatial relationships of each object based on the task requirements and the point cloud accurately.
\\
Returned: \verb|<|\textcolor{codeblue}{\(\pi_{\text{3D}}\)}\verb|>| in text format.
\\
    \textbf{Prompt for unified interface of nav-LLM}\\
\textbf{Prompt of Vision-Language Navigation task:}\\    
Task: Navigate following the instruction: \verb|<|\textcolor{codeblue}{Instruction}\verb|>| \verb|\|n \\
History: Following is the History, which contains the visual information of your previous observation. \verb|<|\textcolor{codeblue}{History}\verb|>| \verb|\|n \\
Observation: Following is the Candidate, which contains several directions you can go to at the current position: \verb|<|\textcolor{codeblue}{Observation}\verb|>| \verb|\|n \\
\(\pi_{\text{3D}}\): Following is the 3D Spatial Description given by a 3D assistant at the current location.\verb|<|\textcolor{codeblue}{\(\pi_{\text{3D}}\)}\verb|>| \verb|\|n \\
Output hint: Select the correct direction from the candidates to go based on the Instruction, Observation and \(\pi_{\text{3D}}\).
\\
\textbf{Prompt of textual question and answer task:}\\    
Task: Please answer questions based on the observation and assistant\verb|\|n \\
Observation: Following is the Observation, which includes multiple images from different locations: \verb|<|\textcolor{codeblue}{Observation}\verb|>| \verb|\|n \\
\(\pi_{\text{3D}}\): Following is the answers to the question given by a 3D assistant.\verb|<|\textcolor{codeblue}{\(\pi_{\text{3D}}\)}\verb|>| \verb|\|n \\
Output hint: Following is the Question you need to answer. Answer the question based on Observation and assistant.\\
\textbf{...}
\end{tcolorbox}
 \label{tab:prompt_table}
\end{table} 

Overview process of our \textit{Navigation Communication Interface} is shown in Table \ref{tab:prompt_table}. Here we show the detailed prompts format used by our Communication Interface providing for various downstream tasks. Here we use the R2R \cite{anderson2018r2r} dataset in the Vision-language navigation (VLN) task and ScanQA dataset in 3D QA as examples. The text in \textcolor{lightpink}{pink means the text data variable in prompt}, \textcolor{lightgreen}{the green means the image data}, \textcolor{lightyellow}{and the yellow means the point cloud data}. Text in \VarSty{codeblue} means the data being exchanged in Communication Interface.
\begin{table}[H]
\label{tab:prompt1}
\begin{tcolorbox} 
\textbf{Prompt format in Communication Interface on VLN task}\\
\\
\VarSty{messages send to 3D-text model} = 
            \{ A chat between a curious user and an artificial intelligence assistant. The assistant gives helpful, detailed, and polite answers to the user's questions. \\
USER: \var{\textcolor{lightyellow}{<point\_start> <point\_patch> <point\_patch>...}\\\textcolor{lightyellow}{<point\_patch ><point\_patch> <point\_end>}}\\
Here is your task: \textcolor{lightpink}{Illustrate the spatial relationships of each object based on the sentence and the point cloud concise and accurately.}\\ ASSISTANT:\}\\
     \VarSty{messages returned to Communication Interface}  =  \{\var{\textcolor{lightpink}{\textbf{<\(\pi_{\text{3D}}\)>}}}\\
     =   A bathroom with blue and white tiles on the walls.\}\\
     \VarSty{messages send to image-text navigation agent}  =  
            \{ Navigate following the instruction.  \\
Instruction: \textcolor{lightpink}{ Walk straight toward the bar with the chairs/stool. Turn left and go straight until you get to three tables with chairs. Turn left and wait near the couch.}\\
Following is the History, which contains the visual information of your previous observation. \\
History: \var{\textcolor{lightgreen}{<img\_patch> <img\_patch>... <img\_patch>}}\\
Following is the Candidate, which contains several directions you can go to at the current position.\\
Candidate: \var{\textcolor{lightgreen}{<img\_patch> <img\_patch>... <img\_patch>}}\\
Following is the 3D Spatial Description of the current location.\\
Spatial Information: \var{\textcolor{lightpink}{\textbf{<\(\pi_{\text{3D}}\)>}}}\\
Output hint: Select the correct direction from the candidates to go based on the Instruction, Observation and Spatial Information.\\
Output: 
      \}
\end{tcolorbox}
\end{table}
\begin{table}[htbp]
\label{tab:prompt2}
\begin{tcolorbox} 
\textbf{Prompt format in Communication Interface on 3D QA task}\\
\\
\VarSty{messages send to 3D-text model} = 
            \{ A chat between a curious user and an artificial intelligence assistant. The assistant gives helpful, detailed, and polite answers to the user's questions. \\
USER: \var{\textcolor{lightyellow}{<point\_start> <point\_patch> <point\_patch>...}\\\textcolor{lightyellow}{<point\_patch ><point\_patch> <point\_end>}}\\
Here is your task: \textcolor{lightpink}{In what part of the room is the long table located?}\\ ASSISTANT:\}\\
     \VarSty{messages returned to Communication Interface}  =  \{\var{\textcolor{lightpink}{\textbf{<\(\pi_{\text{3D}}\)>}}}  =   in center of room \}\\
     \VarSty{messages send to image-text navigation agent}  =  
            \{ Please answer questions based on the 3D Spatial Information and Observation.  \\
Question:  \textcolor{lightpink}{In what part of the room is the long table located?}\\           
Following is the 3D Spatial Information or the answers to the question given by a 3D
assistant.\\
Spatial Information: \var{\textcolor{lightpink}{\textbf{<\(\pi_{\text{3D}}\)>}}}\\
FObservation: Following is the Observation, which includes multiple images from different locations:\\
Observation: \var{\textcolor{lightgreen}{<img\_patch> <img\_patch>... <img\_patch>}}\\
Output hint: Compare the data obtained from 3D assistant and Observation to determine the final answer.\\
Your Answer: \}
\end{tcolorbox}
\end{table}

\begin{table}[p]
\centering
\caption{Details of training time.}
\label{table:training_time}
\resizebox{\textwidth}{!}{
\begin{tabular}{lccccc}
\toprule
 & training of image–text navigation agent&Stage1 & Stage2 & Stage3 & Cross-Modal Belief Alignment \\
\midrule
Cost (1*A100 GPU-Hours) &Refer to NaviLLM \cite{zheng2023towards}&128 & 44 & 24 & 130 \\
\bottomrule
\end{tabular}
}
\end{table}

\begin{table}[tbp]
    \centering
    \caption{Brief Description task example in Stage 1 simple.}
    \begin{tabular}{c|p{8em}|p{8em} }
    \toprule
         Scene  & Instruction  & Ground Truth  \\
         \midrule
                     \raisebox{-1\height}
           { \includegraphics[width=0.2\textwidth]{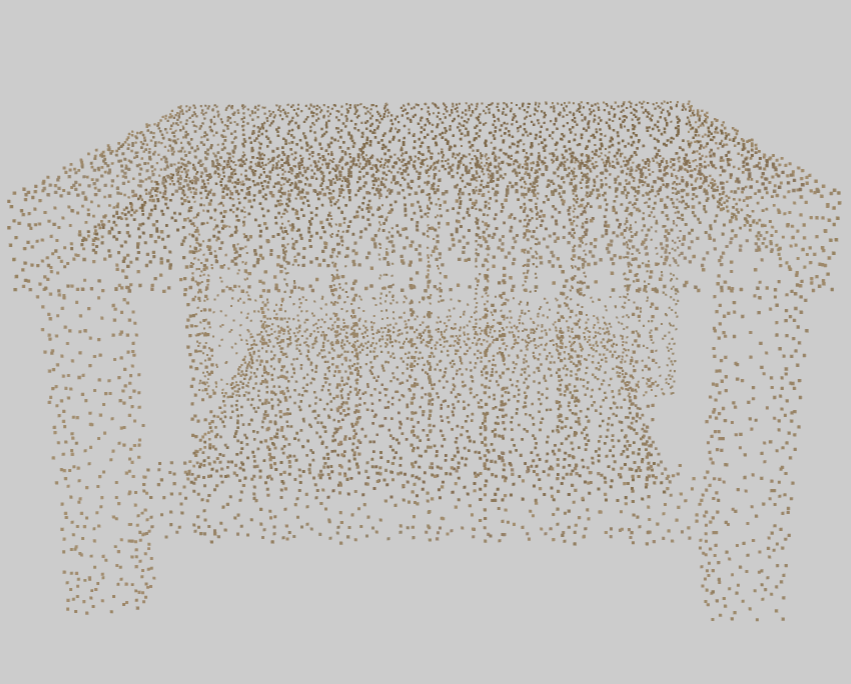} } & Give a quick overview of the object represented by this 3D cloud. &
            A 3D model of a wooden table and bench.\\
         \midrule
            \raisebox{-1\height}
           {\includegraphics[width=0.2\textwidth]{figure/appendix_example/Brief_2.png}}   & Give a brief explanation of the object that this cloud of points forms. & Black coffee cup and saucer 3D model.\\
        \bottomrule
\end{tabular}
\label{tab:append_task_example1}
\end{table}

\begin{table*}[tbhp]
  \centering
  \caption{More visualization examples of point cloud data.}
  \scalebox{0.98}{
    \scriptsize
    \begin{tabular}{l p{14cm}}
      \toprule
      \multicolumn{2}{l}{\bf Some Point Cloud data used in our 3D-text model:} \\
      \midrule
      & 
      \begin{minipage}[c]{0.33\linewidth}
        \centering
        \includegraphics[width=\textwidth,height=0.9\textwidth]{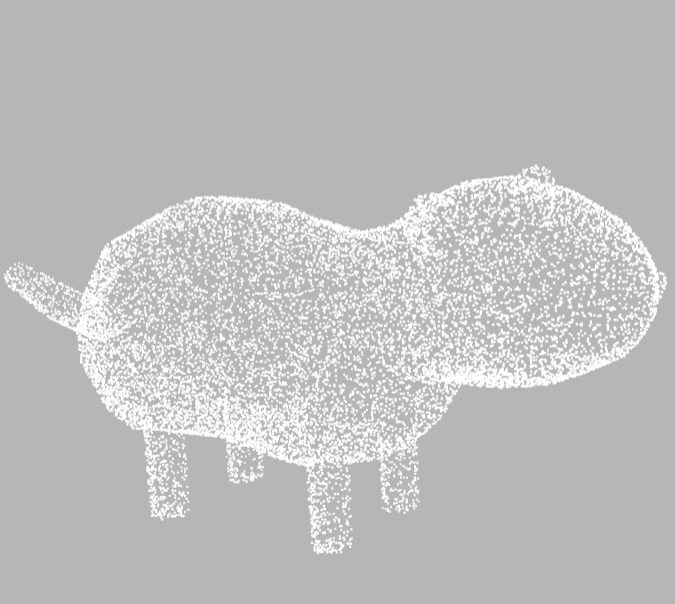} \\
        {\tiny Point cloud from Cap3D.}
      \end{minipage}
      \hfill
      \begin{minipage}[c]{0.33\linewidth}
        \centering
        \includegraphics[width=\textwidth,height=0.9\textwidth]{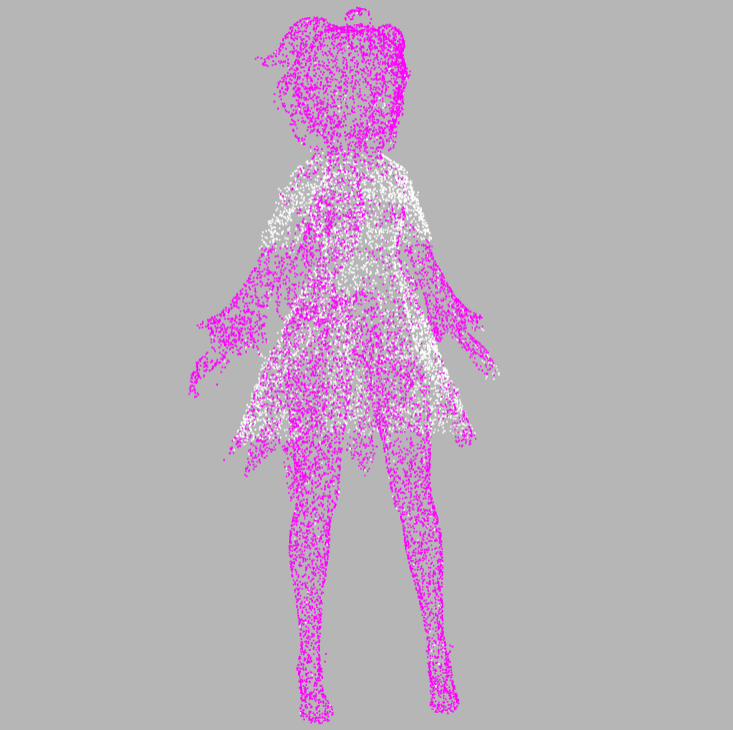} \\
        {\tiny Point cloud from Cap3D.}
      \end{minipage}
      \hfill
      \begin{minipage}[c]{0.33\linewidth}
        \centering
        \includegraphics[width=\textwidth,height=0.9\textwidth]{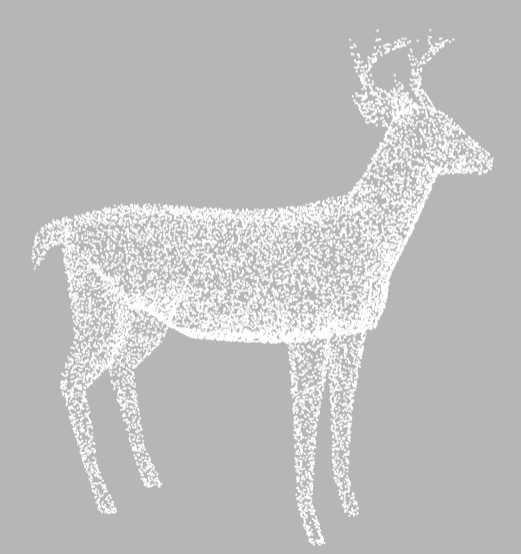} \\
        {\tiny Point cloud from Cap3D.}
      \end{minipage}
      \\ 
      \\
      &
      \begin{minipage}[c]{0.49\linewidth}
        \centering
        \includegraphics[width=\textwidth,height=0.8\textwidth]{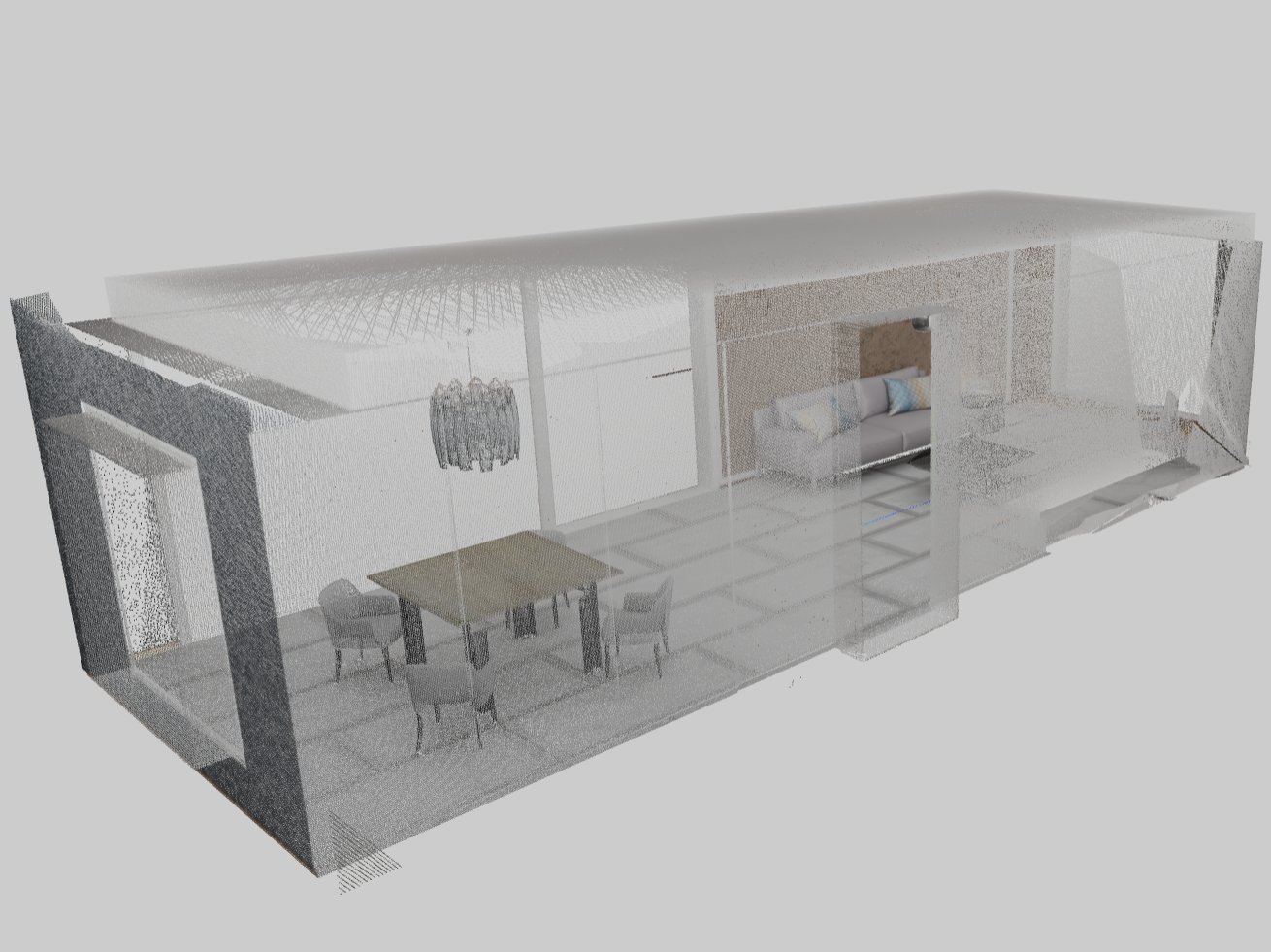} \\
        {\tiny Point cloud from 3D-FRONT.}
      \end{minipage}
      \hfill
      \begin{minipage}[c]{0.49\linewidth}
        \centering
        \includegraphics[width=\textwidth,height=0.8\textwidth]{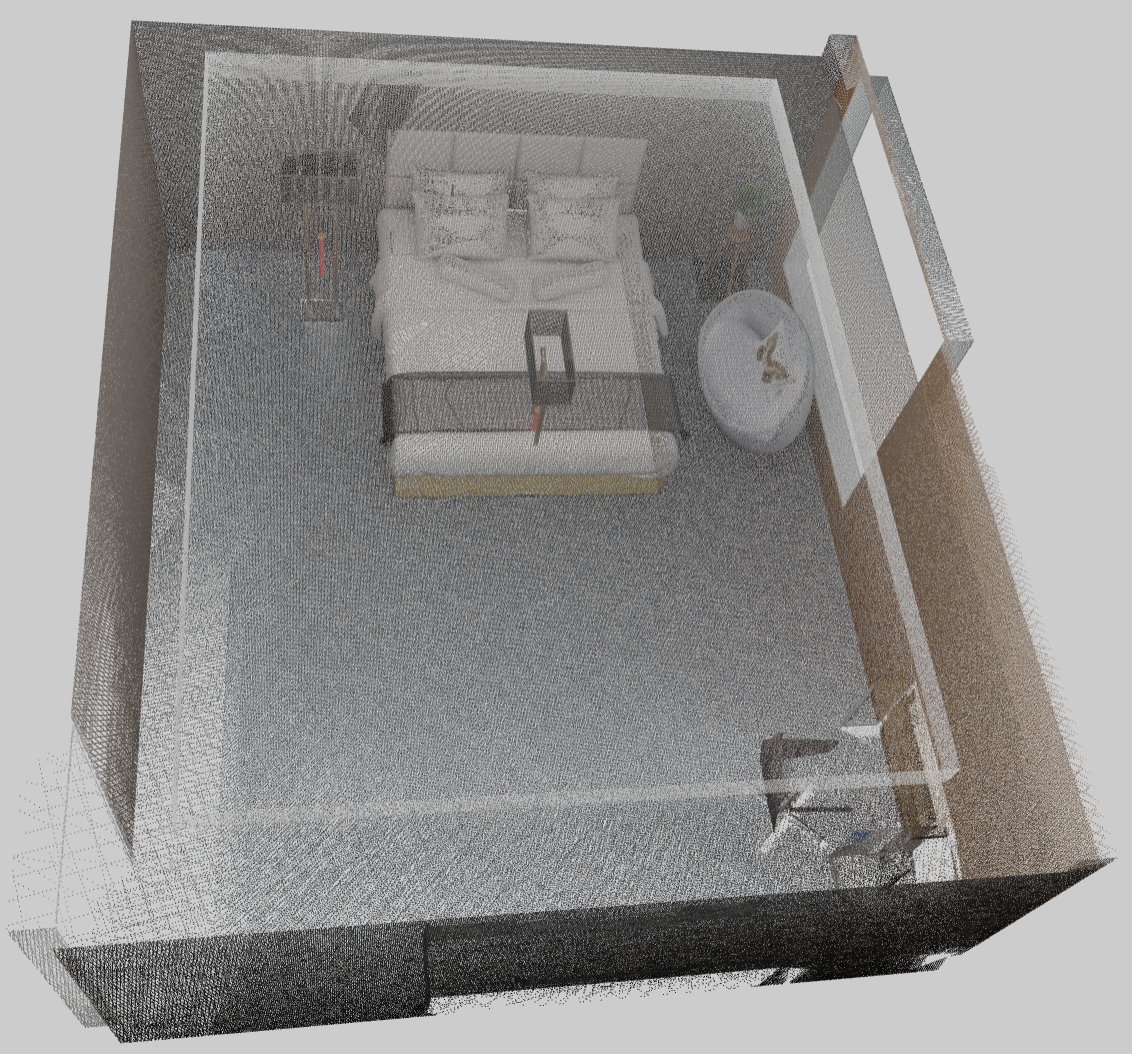} \\
        {\tiny Point cloud from 3D-FRONT.}
      \end{minipage}
      \\ 
      \\
      &
      \begin{minipage}[c]{0.60\linewidth}
        \centering
        \includegraphics[width=\textwidth,height=0.65\textwidth]{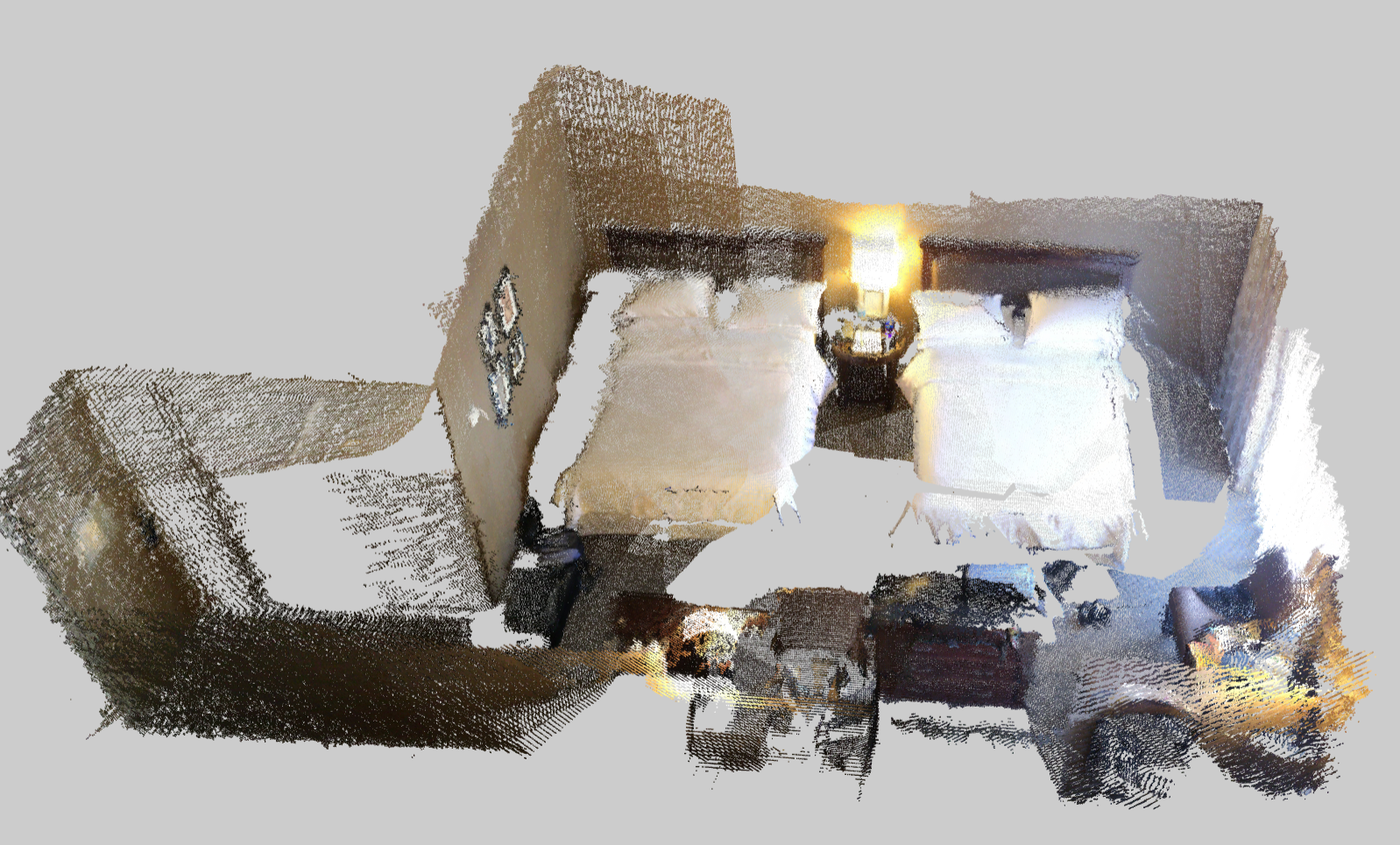} \\
        {\tiny Point cloud from Scannet extracted by us.}
      \end{minipage}
      \hfill
      \begin{minipage}[c]{0.39\linewidth}
        \centering
        \includegraphics[width=0.9\textwidth,height=1.0\textwidth]{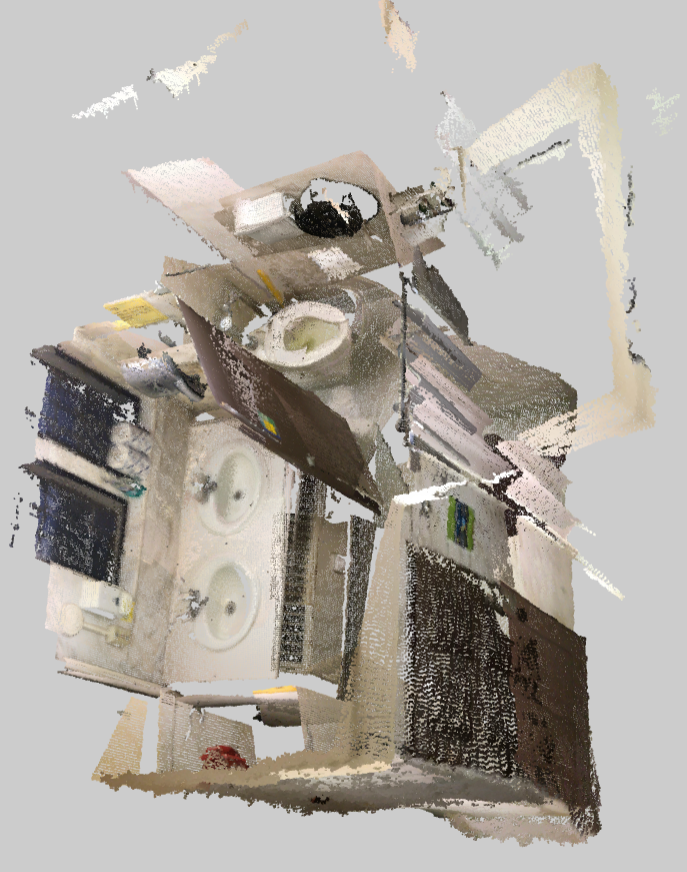} \\
        {\tiny Point cloud from Scannet extracted by us.}
      \end{minipage}
      \\ 
      \\
      &
      \begin{minipage}[c]{0.49\linewidth}
        \centering
        \includegraphics[width=\textwidth,height=0.75\textwidth]{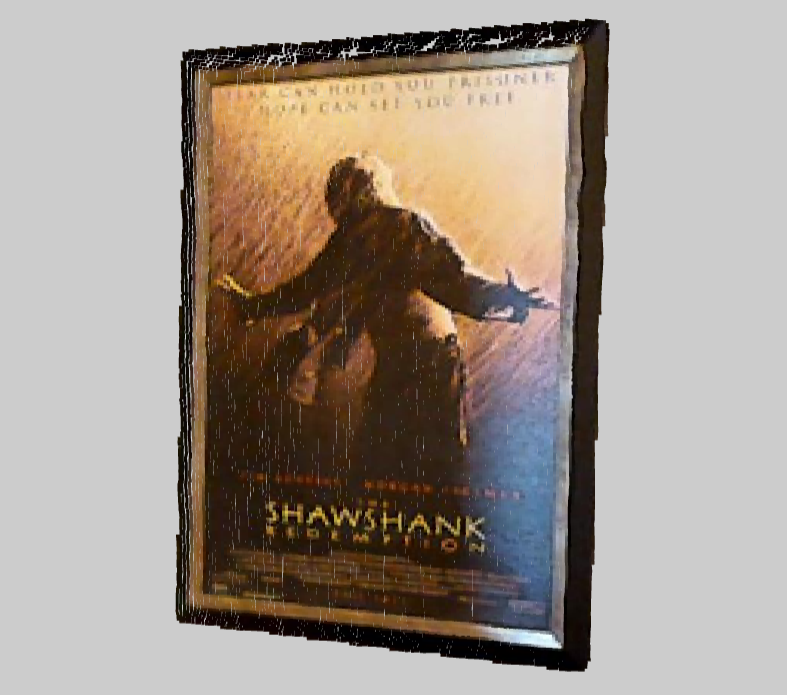} \\
        {\tiny Single object point cloud from R2R extracted by us.}
      \end{minipage}
      \hfill
      \begin{minipage}[c]{0.49\linewidth}
        \centering
        \includegraphics[width=\textwidth,height=0.75\textwidth]{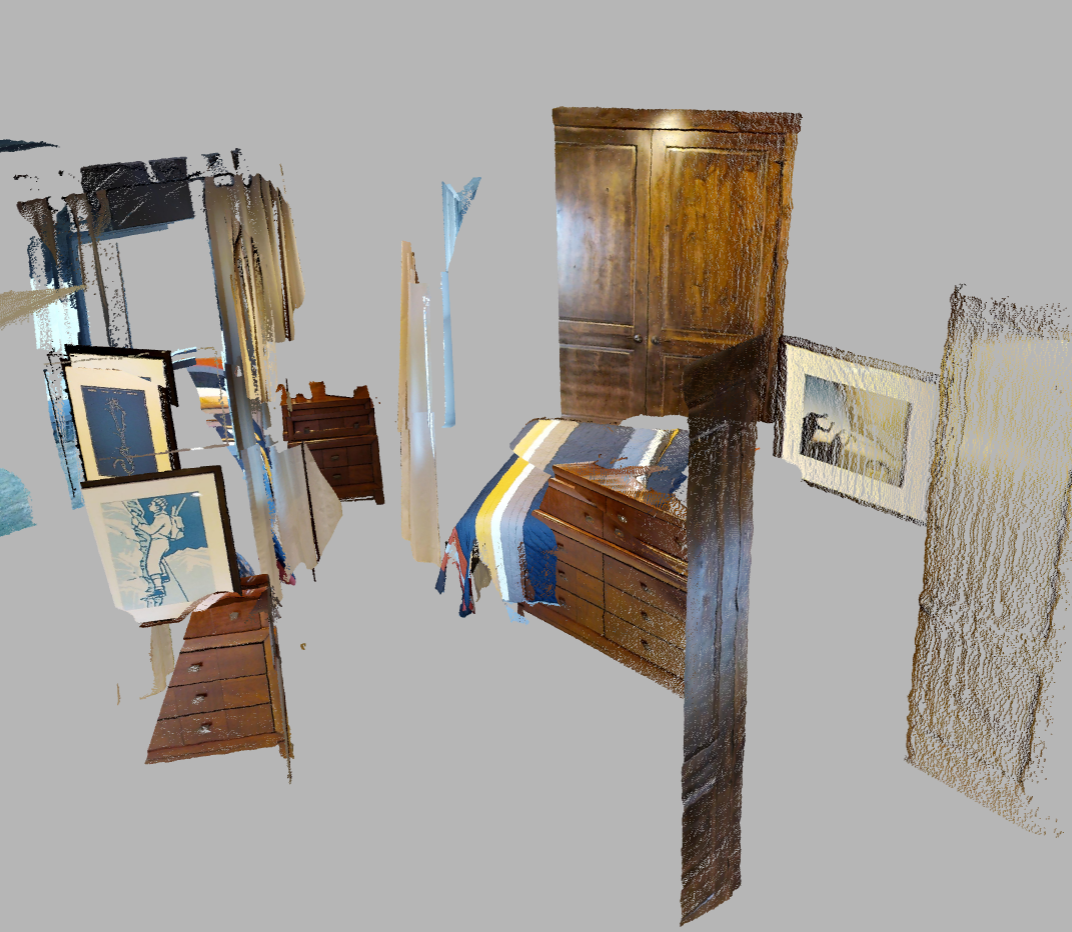} \\
        {\tiny Multi-object point cloud from R2R extracted by us.}
      \end{minipage}
      \\
      \bottomrule
    \end{tabular}
  }
  \label{tab:append_more_visualization_data}
\end{table*}

\begin{table}[tbp]
    \centering
         \caption{Spatial Relation QA in Stage 2 complex.}
    \begin{tabular}{c|p{8em}|p{8em} }
    \toprule
         Scene  & Instruction  & Ground Truth  \\
         \midrule
                     \raisebox{-1\height}
           { \includegraphics[width=0.3\textwidth,,height=0.25\textwidth]{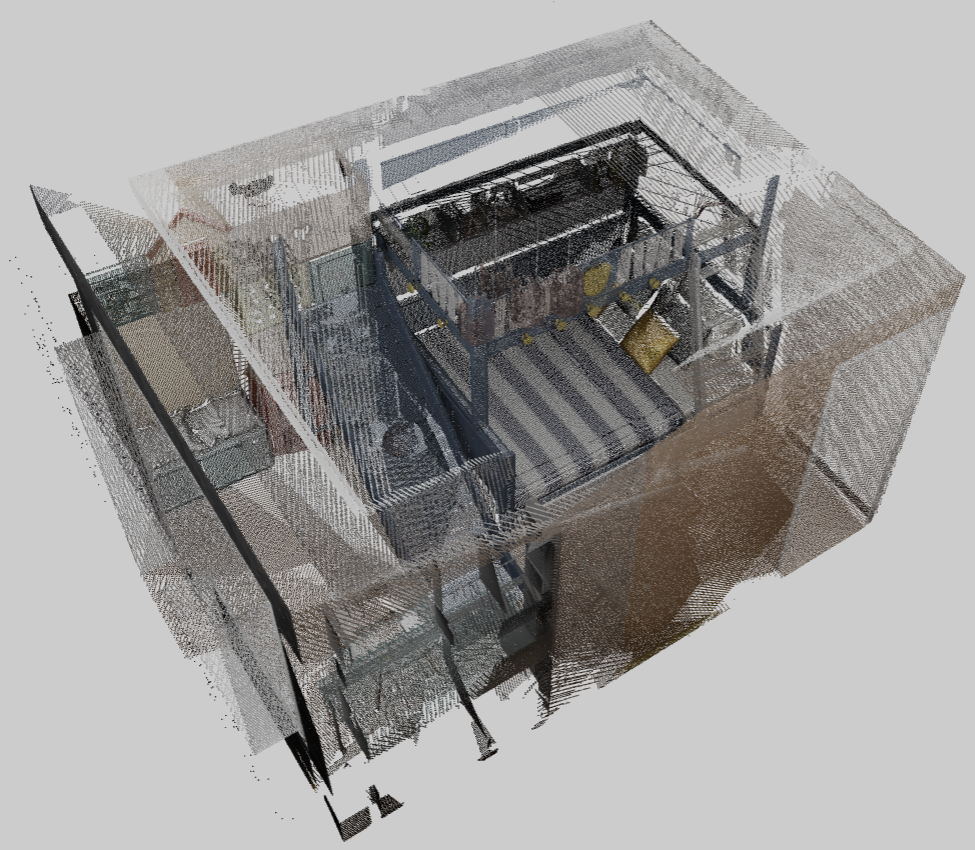} } & What is the spatial relationship between deep blue lounge chair and soft grey canopy ceiling lamp in the KidsRoom? &
            The lounge chair is below the ceiling lamp.\\
         \midrule
            \raisebox{-1\height}
           {\includegraphics[width=0.3\textwidth,,height=0.25\textwidth]{figure/appendix_example/complex_2.png}}   & What is the spatial relationship between classic black tv stand and rich, dark brown three-seat sofa? & The tv stand is far from the three-seat sofa .\\
        \bottomrule
\end{tabular}

\label{tab:append_task_example2}
\end{table}
\begin{table}[tbp]
    \centering
        \caption{Detailed Description in Stage 2 complex.}
              \scriptsize
    \begin{tabular}{c|p{6em}|p{24em} }
    \toprule
         Scene  & Instruction  & Ground Truth  \\
         \midrule

                     \raisebox{-1\height}
           { \includegraphics[width=0.3\textwidth]{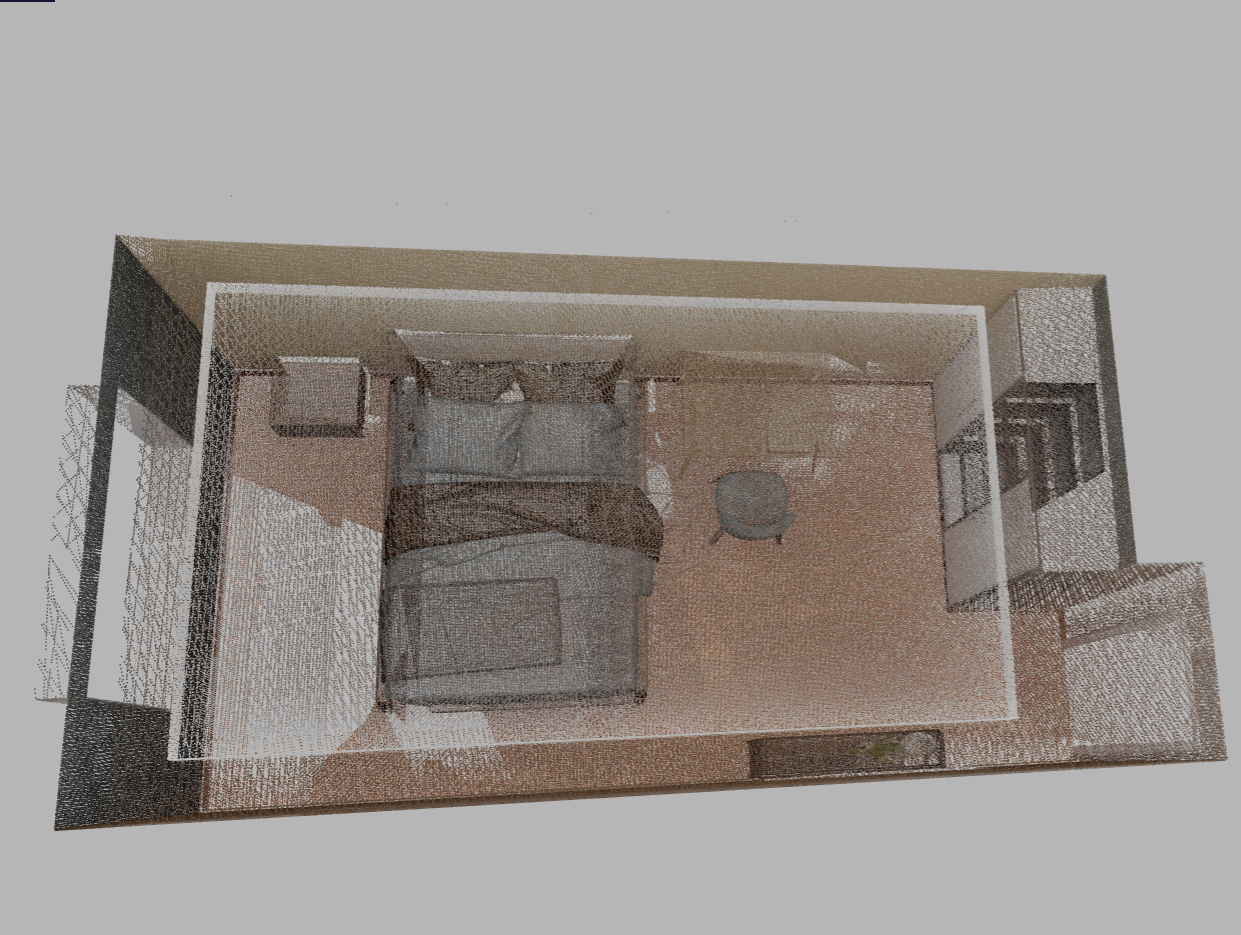} } & Give a thorough rundown of this point cloud. &
            The master bedroom showcases a mix of design styles, from Nordic to Industrial, Japanese to Vintage, all coexisting harmoniously. At the heart of the room lies the king-size bed, flanked by the nightstand, providing a warm and inviting sleep space. Suspended above is the pendant lamp, which casts a modern glow across the room. The wardrobe offers generous storage and contributes an industrial edge with its matte finish. The desk and dining chair are paired together, suggesting a compact area for work or leisure, while the side cabinet provides additional surface space with a retro vibe. This master bedroom is a sanctuary for relaxation, work, and personal care, with each piece of furniture serving a distinct purpose while adding to the room's overall aesthetic.\\
         \midrule
            \raisebox{-1\height}
           {\includegraphics[width=0.3\textwidth]{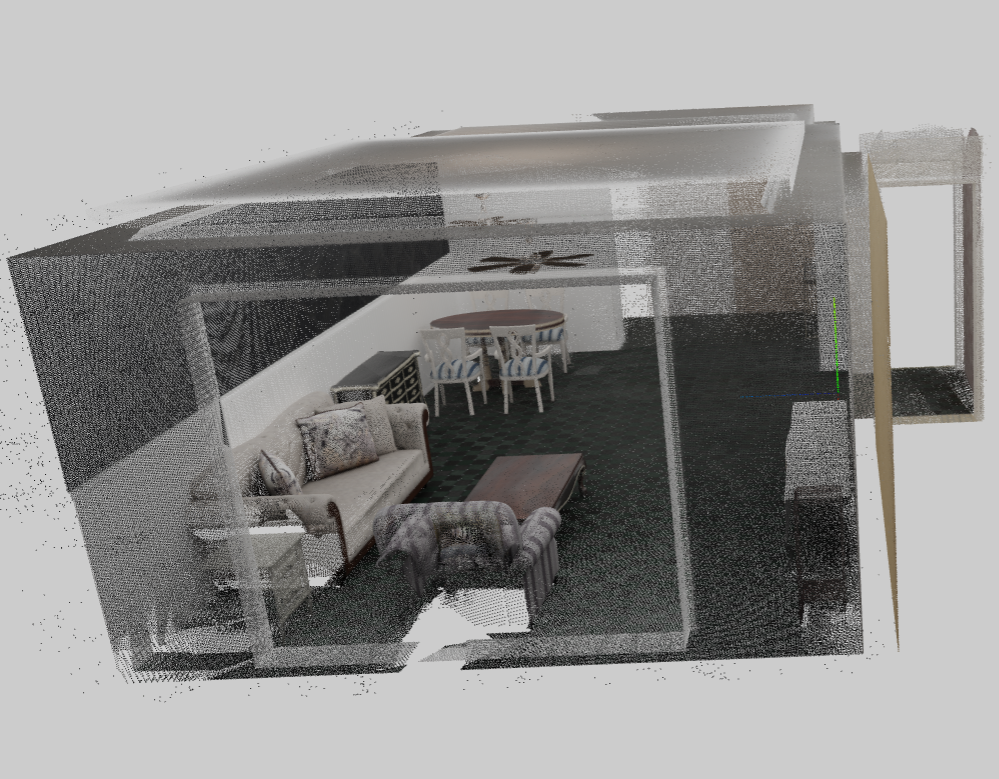}}  &Please detail the specific features of this point cloud. & The living room is a harmonious blend of various design styles, creating a multifunctional space for relaxation and dining. The corner table with its Japanese style, is placed subtly in the room, offering a spot for decorative items or small necessities. The tv stand exudes the elegance of Southeast Asia and serves as a central entertainment hub. In front of it, the three-seat sofa invites relaxation with its minimalist light grey upholstery, while the coffee table sits conveniently in front, echoing European classic design. The armchair with its American Country style provides additional seating next to the coffee table, perfect for socializing or enjoying a quiet read. The side cabinet brings an industrial touch to the room, offering storage and display space. This room is designed for both leisure and dining, with a layout that encourages flow and interaction.\\
        \bottomrule
\end{tabular}
\label{tab:append_task_example3}
\end{table}
\begin{table}[tbp]
    \centering
         \caption{Brief Description in Stage 3 real-world.}
    \begin{tabular}{c|p{8em}|p{8em} }
    \toprule
         Scene  & Instruction  & Ground Truth  \\
         \midrule
                     \raisebox{-1\height}
           { \includegraphics[width=0.3\textwidth]{figure/qualitative_results_r2r_v2.jpg} } & I'm interested in this, can you explain? &
           A living room filled with brown leather furniture and pictures on the wall.\\
         \midrule
            \raisebox{-1\height}
           {\includegraphics[width=0.3\textwidth]{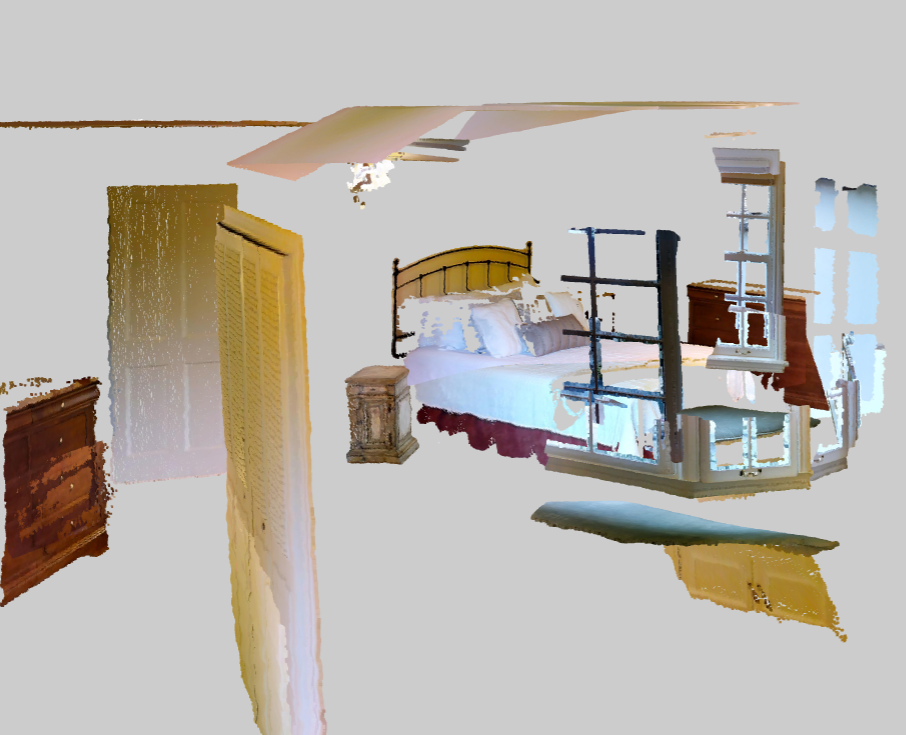}}   & Illustrate the  the point cloud concise and accurately. & A bedroom with a bed, dresser and ceiling fan in it.\\
        \bottomrule
\end{tabular}

\label{tab:append_task_example4}
\end{table}

\begin{table}[tbp]
    \centering
         \caption{3D QA in Stage 3 real-world.}
    \begin{tabular}{c|p{8em}|p{8em} }
    \toprule
         Scene  & Instruction  & Ground Truth  \\
         \midrule
                     \raisebox{-1\height}
           { \includegraphics[width=0.3\textwidth]{figure/qualitative_results_scannet_v2.jpg} } & On what side of the cabinets is the microwave located? &
           right side.\\
         \midrule
            \raisebox{-1\height}
           {\includegraphics[width=0.3\textwidth]{figure/appendix_example/4_1.png}}   & What color leather sofa is in the room? & brown\\
        \bottomrule
\end{tabular}

\label{tab:append_task_example5}
\end{table}

\begin{table*}[btp!]
    \centering
\begin{minipage}{0.75\linewidth}
\centering
\captionsetup{width=0.95\textwidth}
\caption{\footnotesize Top Teams on the ScanQA Test without Objects split Leaderboard Until May 2025.}
\resizebox{\textwidth}{!}{
\begin{tabular}{l|l|cccccc} \hline
Rank&Team&\textbf{EM}$\uparrow$&\textbf{BLEU-1}$\uparrow$ & \textbf{BLEU-4}$\uparrow$ & \textbf{ROUGE}$\uparrow$ & \textbf{METEOR}$\uparrow$ & \textbf{CIDEr}$\uparrow$ \\ \toprule
1&Rabbit & 35.6&47.22&15.45&49.53&20.05&100.44 \\
2&mare-ephemeral & 32.70	&	42.53	&18.52	&45.40&	18.45	&91.50 \\
3&OldStone&30.80	&	34.60	&0.00	&41.32&	15.66	&78.72\\ 
4&MilkDragon &30.17	&	33.45	&19.07	&40.71	&15.26	&77.49\\
\rowcolor{Gray!20}
5&CoNav	&25.11	&	37.51&	11.43&	37.64	&15.21	&73.51\\
\bottomrule
\end{tabular}
}
\label{table:scanqa_sota}
\end{minipage}
\end{table*}

\begin{table*}[tbp]
\centering
\caption{Hyperparameters for All Stages in training 3D-text model.} 
\label{tab:append_param_all}
\resizebox{\textwidth}{!}{
\begin{tabular}{c|ccc}
\toprule
\textbf{Hyperparameter}         & \textbf{Stage 1 Value}     & \textbf{Stage 2 Value}                        & \textbf{Stage 3 Value}      \\ \midrule
Learning rate                   & $2\times 10^{-5}$          & $1\times 10^{-6}$                             & $2\times 10^{-5}$            \\
Parallel strategy               & DDP                        & FSDP (full shard auto wrap)                   & DDP                          \\
Batch size total                & 16                         & 4                                             & 8                           \\
Epochs                          & 3                          & 1                                             & 3                            \\
\multirow{2}{*}{Task Type   }                  & \multirow{2}{*}{ Brief Description }        &  Spatial Relation QA,  & 3D QA,    \\ 
& &  Detailed Description &Brief Description\\
\midrule
Optimizer                       & \multicolumn{3}{c}{AdamW}                                                                             \\
Warmup ratio                    & \multicolumn{3}{c}{0.03}                                                                              \\
LR scheduler type               & \multicolumn{3}{c}{cosine}                                                                           \\
Number of workers               & \multicolumn{3}{c}{4}                                                                                \\
Type of GPUs                    & \multicolumn{3}{c}{NVIDIA A100 (80G with NvLink)}                                                     \\
Number of GPUs                  & \multicolumn{3}{c}{8}                                                                                \\
Accumulate gradient batches     & \multicolumn{3}{c}{1}                                                                                \\
Training precision              & \multicolumn{3}{c}{bfloat16}                                                                         \\ \bottomrule
\end{tabular}%
}
\end{table*}

\begin{table*}[p]
\centering
\captionof{table}{Hyperparameters for the Cross-Modal Belief Alignment.} 
\resizebox{0.6\linewidth}{!}{
\label{tab:append_param_nav}
\begin{tabular}{@{}ll@{}}
\toprule
\textbf{Hyperparameter}  & \textbf{Value}   \\ \midrule
Optimizer                & AdamW     \\
Learning rate            & $1\times 10^{-5}$ \\
warmup ratio            & 0.05     \\
lr scheduler type&cosine\\
Number of workers        & 4         \\
Type of GPUs             & NVIDIA A100   (80G with NvLink)    \\
Number of GPUs           & 8 \\
Accumulate gradient batches & 1      \\
Batch size total   & 4    \\ 
Training precision       & bfloat16      \\
\multirow{2}{*}{Task Type} &3D QA, Navigation,  \\
& Path Summarization\\

\bottomrule
\end{tabular}
}
\end{table*}

\begin{table*}[p]
\centering
\begin{minipage}{0.35\linewidth}
\centering
\captionof{table}{Hyperparameters for inference.}
\resizebox{\linewidth}{!}{
\label{tab:append_param_inf}
\begin{tabular}{@{}ll@{}}
\toprule
\textbf{Hyperparameters}  & \textbf{Value}   \\ \midrule
\multicolumn{2}{c}{In 3D-text model }\\ \midrule
Max new tokens    & 64   \\
Top $p$                   & 0.95   \\ 
Do Sample          & True   \\
Temperature              & 0.2   \\ \midrule
\multicolumn{2}{c}{In image-text navigation agent }\\ \midrule
Max new tokens    & 20   \\
Top $p$                   & 0.95   \\ 
Do Sample          & True   \\
Temperature              & 0.1   \\
\bottomrule
\end{tabular}
}
\end{minipage}
\end{table*}

\clearpage

\clearpage 
\end{document}